\documentclass[lettersize,journal]{IEEEtran}
\usepackage{amsmath,amsfonts}
\usepackage{algorithmic}
\usepackage{algorithm}
\usepackage{array}
\usepackage[caption=false,font=normalsize,labelfont=sf,textfont=sf]{subfig}
\usepackage{textcomp}
\usepackage{stfloats}
\usepackage{url}
\usepackage{verbatim}
\usepackage{graphicx}
\usepackage{cite}
\usepackage{multirow}
\usepackage{booktabs}
\usepackage{xcolor}
\usepackage{colortbl}
\usepackage{orcidlink}
\usepackage{bbding}
\usepackage{makecell}
\usepackage{colortbl}
\usepackage{color}
\usepackage[switch]{lineno}
\usepackage{pifont}       
\usepackage{bbding}  

\hypersetup{%
    pdfborder = {0 0 0}
}
\definecolor{mygray}{gray}{.9}
\definecolor{myblue}{RGB}{93,80,180}
\definecolor{mygreen}{RGB}{93,173,85}
\begin{document}

\title{STeInFormer: Spatial-Temporal Interaction Transformer Architecture for Remote Sensing Change Detection}

\author{Xiaowen Ma~\orcidlink{0000-0001-5031-2641}, 
Zhenkai Wu \orcidlink{0009-0000-0613-0584}, 
Mengting Ma~\orcidlink{0000-0002-6897-3576},
Mengjiao Zhao \orcidlink{0009-0006-1814-2404},
Fan Yang~\orcidlink{0009-0002-1292-0894},
Zhenhong Du \orcidlink{0000-0001-9449-0415}, 
and Wei Zhang~\orcidlink{0000-0002-4424-079X} \vspace{.5em} \\

\thanks{This work was supported by the Public Welfare Science and Technology Plan of Ningbo City (2022S125) and the Key Research and Development Plan of Zhejiang Province (2021C01031). \emph{(Corresponding author: Wei Zhang.)}}

\thanks{Xiaowen Ma, Zhenkai Wu, Fan Yang and Mengjiao Zhao are with the School of Software Technology, Zhejiang University, Hangzhou 310027, China. (e-mail: xw.ma@zju.edu.cn)}

\thanks{Mengting Ma is with the School of Computer Science and Technology, Zhejiang University, Hangzhou 310027, China.}

\thanks{Zhenhong Du is with the School of Earth Sciences, Zhejiang University, Hangzhou 310027, China.}

\thanks{Wei Zhang is with the School of Software Technology, Zhejiang University, Hangzhou 310027, China, and also with the Innovation Center of Yangtze River Delta, Zhejiang University, Jiaxing Zhejiang, 314103, China (e-mail: cstzhangwei@zju.edu.cn).}

}

\markboth{Submit to IEEE Journal of Selected Topics in Applied Earth Observations and Remote Sensing}%
{Shell \MakeLowercase{\textit{et al.}}: A Sample Article Using IEEEtran.cls for IEEE Journals}

\maketitle

\begin{abstract}
Convolutional neural networks and attention mechanisms have greatly benefited remote sensing change detection (RSCD) because of their outstanding discriminative ability. Existent RSCD methods often follow a paradigm of using a non-interactive Siamese neural network for multi-temporal feature extraction and change detection heads for feature fusion and change representation. However, this paradigm lacks the contemplation of the characteristics of RSCD in temporal and spatial dimensions, and causes the drawback on spatial-temporal interaction that hinders high-quality feature extraction. To address this problem, we present STeInFormer, a spatial-temporal interaction Transformer architecture for multi-temporal feature extraction, which is the first general backbone network specifically designed for RSCD. In addition, we propose a parameter-free multi-frequency token mixer to integrate frequency-domain features that provide spectral information for RSCD. Experimental results on three datasets validate the effectiveness of the proposed method, which can outperform the state-of-the-art methods and achieve the most satisfactory efficiency-accuracy trade-off. Code is available at \textbf{\url{https://github.com/xwmaxwma/rschange}}.
\end{abstract}


\maketitle

\section{Introduction}
Remote sensing imagery has been growing drastically along the development of earth observation technology \cite{cui2022unsupervised,cui2024real,cui2024superpixel}, encouraging the communities of earth sciences and remote sensing to adopt deep learning (DL) techniques for relevant tasks~\cite{rstask,CMSCGC,SSGCC}. Remote sensing change detection (RSCD) focuses on comparing two or more images taken at different times in the same region for quantitative and qualitative assessment of changes in geographic entities and environmental factors, usually in a multi-scale and multi-temporal context. It is of high scientific and practical importance and serves a wide range of goals, such as environmental monitoring~\cite{monitor}, urban planning~\cite{urban}, disaster assessment~\cite{disaster}, and land use~\cite{land,land2,crossmatch}.

\begin{figure}[t]
	\centering
	\includegraphics[width=0.48\textwidth]{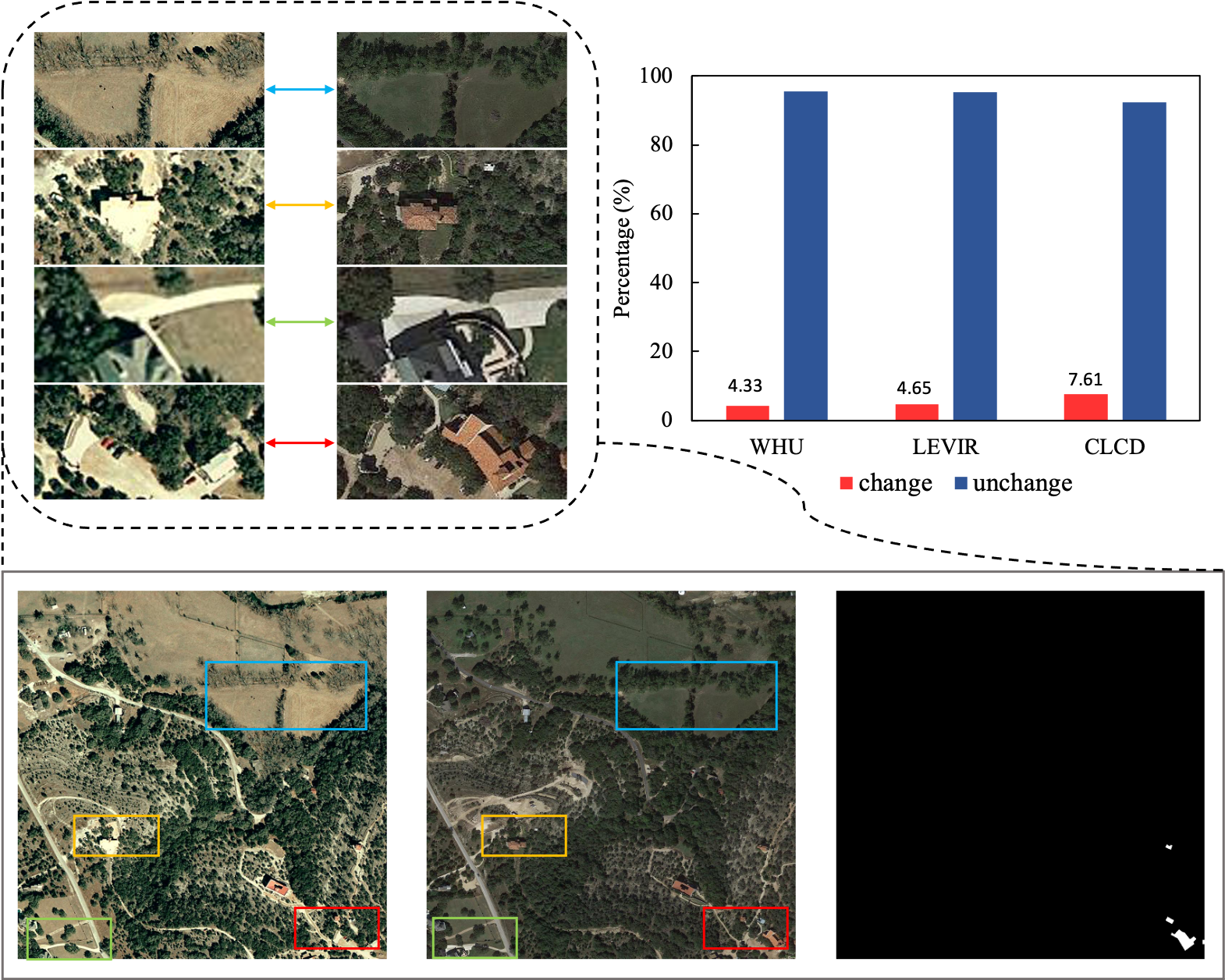}
	\caption{Visualization of two challenges in RSCD: frequent non-interest changes and the requirement for high spatial detail. Example changes of interest changes (red box) and non-interest changes (blue box: non-interest objects; orange box: illumination variations; green box: registration errors) are shown in the lower bi-temporal images. The upper-right chart illustrates the imbalanced distributions of the number of changed pixels and that of the non-changed on three datasets.}
	\label{fig:intro}
\end{figure}

RSCD can be regarded as a binary semantic segmentation problem, which assigns to each pixel a binary label indicating if the \emph{object of interest} in the corresponding area has changed or not \cite{chen2023self,chen2024learning,qianmaskfactory}. In practice, a significant challenge occurs to RSCD due to frequent \emph{non-interest changes} caused by seasonal illumination variations, unrelated movements, or even differences in sensors and imaging conditions. Besides, the sizes of changed areas could be much smaller than the unchanged in a target region over a certain time span, requiring rich spatial details to detect. The afore-mentioned two challenges are visualized in Fig.~\ref{fig:intro} and should be carefully addressed for RSCD.


Most of traditional RSCD methods are based on algebra~\cite{cha,bi} and transformation~\cite{pca,cva}. Despite being simple to implement, these methods rely on hand-crafted features, and experience high computational complexity as well as noise sensitivity. The recent burgeoning of deep learning techniques \cite{chen2024tokenunify,chen2024bimcv, sunprogram}, especially convolutional neural networks (CNNs)~\cite{cnn}, has greatly facilitated the development of RSCD, given their outstanding capability of non-linear fitting for extracting high-quality discriminative features.~\cite{fc-siam} introduces the Siamese neural network to RSCD, which extracts bi-temporal features followed by a change detection head using splicing or summation. This paradigm can be further implemented with a weight-sharing tandem classification network~\cite{vgg,resnet} as the backbone to improve the change detection head. For example,~\cite{ifnet,stanet,dtcdscn} enhance feature representation based on spatial attention and channel attention;~\cite{tfigr} optimizes splicing or phase-subtraction to refine temporal feature interaction. However, a large gap still exists on semantic information and spatial detail among multi-level features obtained by tandem classification networks, while the redundancy in the channels of deep features leads to a huge computational cost. In addition, the U-shaped architecture adopted by~\cite{snunet} can superimpose and fuse features at different levels improving the method's capability to distinguish changed and unchanged areas, but comprises dense connectivity also causing the afore-mentioned computational problem.

More recent studies have adopted Transformer~\cite{transformer} for RSCD to bypass the limitations of CNNs regarding fixed perceptual fields and weakly-captured long-range dependence. For example,~\cite{swinsunet} proposes a purely Transformer-based RSCD network using the SwinTransformer~\cite{swin};~\cite{changeformer} extracts coarse-grained and fine-grained features from bi-temporal images by constructing a pair of Siamese neural networks with a layered Transformer encoder;~\cite{bit} employs Transformer encoders to model the context in a compact token-based space-time, where the learned context-rich tokens are fed to the pixel space for refining the original features by the Transformer decoder. However, these methods also follow the tandem design of classification networks~\cite{vit, swin}, and attention mechanisms require high computation (i.e., squared time complexity and space complexity).

Upon the observation on the above-discussed methods, which follow a paradigm of non-interactive Siamese neural network and change detection head but hardly consider RSCD's characteristics, we assume that integrating cross-temporal and cross-spatial interactions of features in feature extraction can benefit the improvement of performance for RSCD. Following this assumption, we propose \emph{STeInFormer}, a novel RSCD method with a spatial-temporal interaction Transformer. The proposed method consists of cross-temporal interactors (CTIs) and cross-spatial interactors (CSIs). Specifically, the CTI adopts the gating mechanism to emphasize changes of interest while suppressing non-interest changes in feature extraction; and the CSI serves as an encoding stage based on the U-shaped architecture to integrate semantic and detail information for a more robust feature representation. Our STeInFormer is the first architecture entirely designed for RSCD, whose capability has been validated by extensive experiments to serve as a generic backbone for change detection tasks. In addition, it is noteworthy that previous methods solve the problem solely from the perspective of spatial domain, whereas studies in other topics have demonstrated the potential of considering frequency domain in image processing~\cite{ff1, ff2, fcanet}. Therefore, we devise a parameter-free multi-frequency mixer based on the discrete cosine transform (DCT), which combines prior frequency values to integrate token information and has linear complexity to enable extreme lightweight.

The contributions of our work can be summarized as follows:
\begin{itemize}
\item We design a novel paradigm to accomplish temporal and spatial interactions in feature extraction for robust feature representation, following which we propose the first architecture as a generic backbone specifically for RSCD.
\item We consider the characteristics of RSCD and introduce the gating mechanism and the U-shaped architecture to enable cross-temporal and cross-spatial interactions, respectively.
\item We devise a multi-frequency mixer that for the first time serves RSCD from the perspective of frequency domain and has linear complexity using prior frequency values.
\item Extensive experiments on three datasets suggest that the proposed method can outperform other state-of-the-art methods, while achieving a better efficiency-accuracy trade-off.
\end{itemize}

\section{Related Works}
\subsection{Traditional RCSD Methods}
Early methods for RSCD can be divided into the algebra-based, the transformation-based and the classification-based. Algebraic-based methods adopt image difference~\cite{cha}, image ratio~\cite{bi} and image regression~\cite{hui} to obtain a difference map and select appropriate thresholds to identify the areas of change. Transformation-based methods, such as Principal Component Analysis~\cite{pca}, Change Vector Analysis~\cite{cva}, and Tasseled Cap Transformation~\cite{cap}, use a spatial mapping to highlight changes of interest. Classification-based methods rely on traditional machine learning algorithms, such as support vector machines (SVMs)~\cite{svm} and random forests~\cite{rand}, to discriminate changed pixels. However, the above-mentioned methods depend greatly on the quality of empirically designed and hand-crafted features and thus easily result in unsatisfactory results on high-resolution remote sensing imagery.

\begin{figure*}[t]
	\centering
	\includegraphics[width=0.98\textwidth]{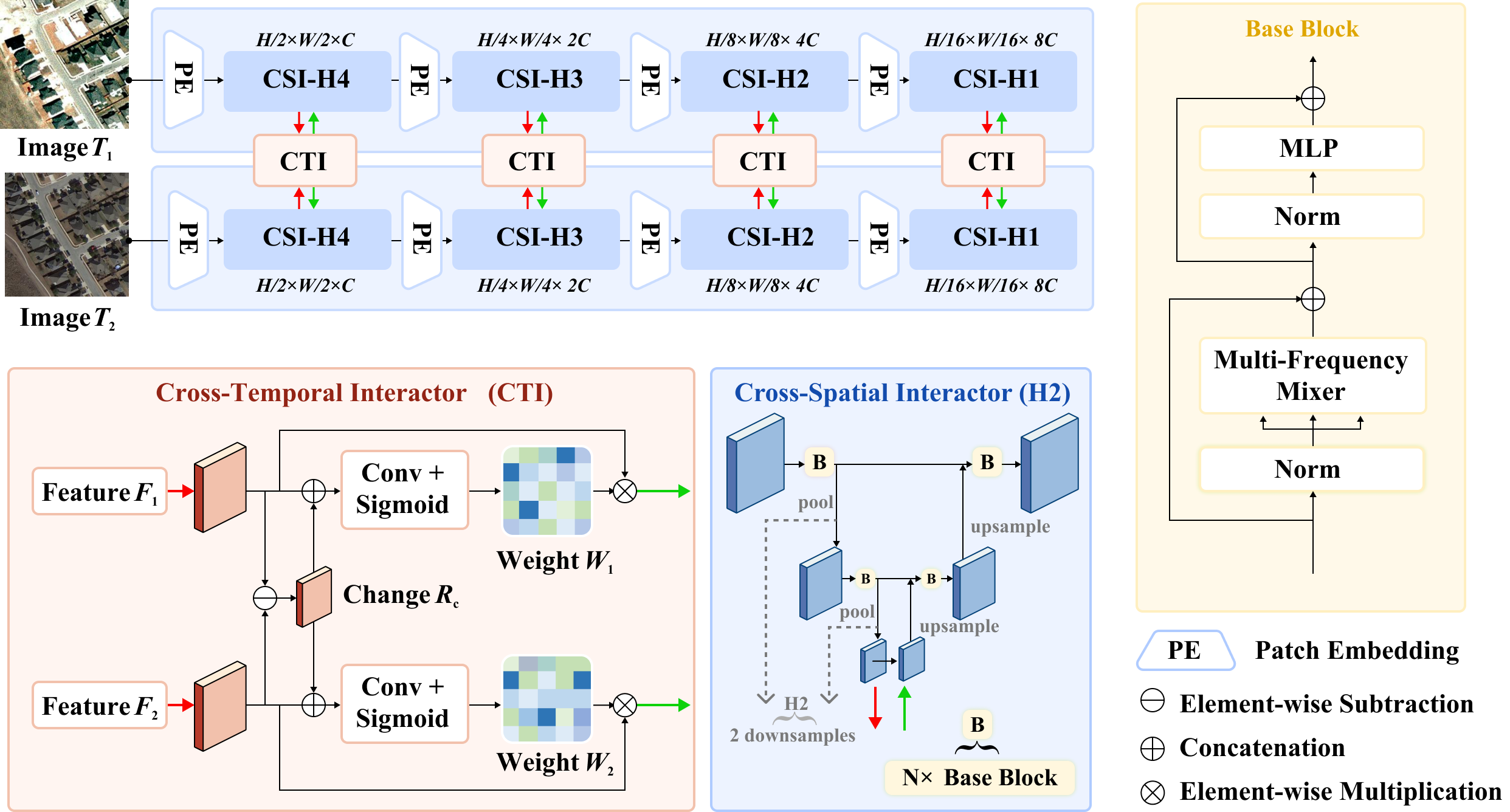}
	\caption{Architecture of the STeInFormer. Given as input bi-temporal images, multi-scale features are extracted by each CSI, which is U-shaped and relies on base blocks, to fuse the semantic information of high-level features and the spatial detail of low-level features. The deepest features of each CSI are fed to the corresponding CTI for cross-temporal interaction. H-i denotes that i times downsampling is implemented by CSI to ensure that the resolution of the feature maps input to the CTI module are all 1/32 of the original images. The STeInFormer outputs bi-temporal features at four scales. }
	\label{fig:whole}
\end{figure*}

\subsection{Deep Learning-Based RCSD Methods}
Most of deep learning-based RCSD methods can be categorized into the CNN-based and the Transformer-based. Restricted by fixed receptive fields, CNN-based methods contribute mainly to contextual modeling enhancing the capability to distinguish changes from bi-temporal images, which includes spatial context modeling (e.g., dilated convolutional strategies~\cite{spatial}) and relational context modeling (e.g., attention mechanisms~\cite{ifnet, stanet,dtcdscn, dminet}). In addition, some studies~\cite{deeply1, deeply2, snunet} exploit multi-level feature fusion to integrate high-level features with sufficient semantic information, and low-level features with fine spatial details, so as to obtain a better representation of change. However, these methods can hardly model the global context at the early stage of the network and lack spatio-temporal interaction for more robust and discriminative bi-temporal features.

Existing transformer-based RSCD approaches are mainly based on multi-head attention mixers to capture long-range contextual information. For example,~\cite{mstdsnet, swinsunet} employ the SwinTransformer~\cite{swin} to aggregate multi-level features;~\cite{changeformer} follows~\cite{
segformer} to devise a hierarchical Transformer encoder;~\cite{bit} proposes a hybrid structure to obtain semantic tokens with a CNN and model the context in the spatio-temporal domain with the Transformer. Although these methods demonstrate efficacy in capturing the global context, they still hardly consider spatio-temporal interaction \cite{nie2024imputeformer,he2024geolocation} and thus are inadequately capable to address RSCD's challenges on frequent non-interest changes and high spatial details. In addition, the computation of the attention mechanism requires squared time complexity and space complexity, resulting in a high computational cost.

\subsection{Transformer-based Methods for Vision}
Transformers~\cite{transformer} are initially proposed for natural language processing tasks, and later~\cite{vit} introduces to the field of computer vision achieving considerable successes. Subsequent works focus on optimizing attention modules, such as shifting windows~\cite{swin}, hierarchical patch embedding~\cite{hivit}, relative position encoding~\cite{relative}, improving attention mapping~\cite{refiner, sun2024ultrahighresolutionsegmentationboundaryenhanced, cheng2024sptsequenceprompttransformer}, and merging convolutions~\cite{cvt,cmt, convit}, to improve token mixing. Recently, studies adopting MLPs~\cite{mlp, mlps, resmlp} as token mixers have achieved competitive performances, to replace attention modules. However, high computation still limits Transformer's widespread application. To address this issue, ~\cite{poolormer} uses average pooling for token mixing greatly reducing the number of parameters as well as computational effort, whereas the obtained representations are over-homogeneous.~\cite{fnet} exploits the Fourier transform (FT) to mix tokens with the real part retained, which involves complex operations and has high time complexity. The DCT's computational efficiency and effectiveness in capturing local image features motivate us to design a multi-frequency mixer that combines the DCT and priori frequency values.

\section{Method}
\label{sec:method}
In this section, we provide an overview of the proposed method regarding its architecture (see Sec.~\ref{sec:method:arch}), including cross-temporal interactors (Sec.~\ref{sec:method:cti}) and cross-spatial interactors (Sec.~\ref{sec:method:csi}). Besides, we introduce the multi-frequency mixer (Sec.~\ref{sec:method:mfm}), which serves as a token mixing module, and the loss function (Sec.~\ref{sec:method:loss}).

\subsection{Architecture}
\label{sec:method:arch}
As shown in Fig.~\ref{fig:whole}, the architecture of our STeInFormer is mainly composed of cross-spatial interactors (CSIs) and cross-temporal interactors (CTIs). First, the input bi-temporal images $\mathcal{T}_1$ and $\mathcal{T}_2$ are transformed into embedded tokens by patch embedding (PE) blocks. These tokens are then fed into CSIs for feature extraction. In particular, the deepest features of each CSI as an encoding stage are extracted and given to a corresponding CTI for cross-temporal interaction, which produces the temporal disparity-enhanced features returned to the CSI, where spatial details are recovered after multi-level upsampling and jump connections. Therefore, our STeInFormer conducts cross-temporal and cross-spatial interaction of bi-temporal features at each encoding stage.

\begin{figure*}[t]
	\centering
	\includegraphics[width=0.99\textwidth]{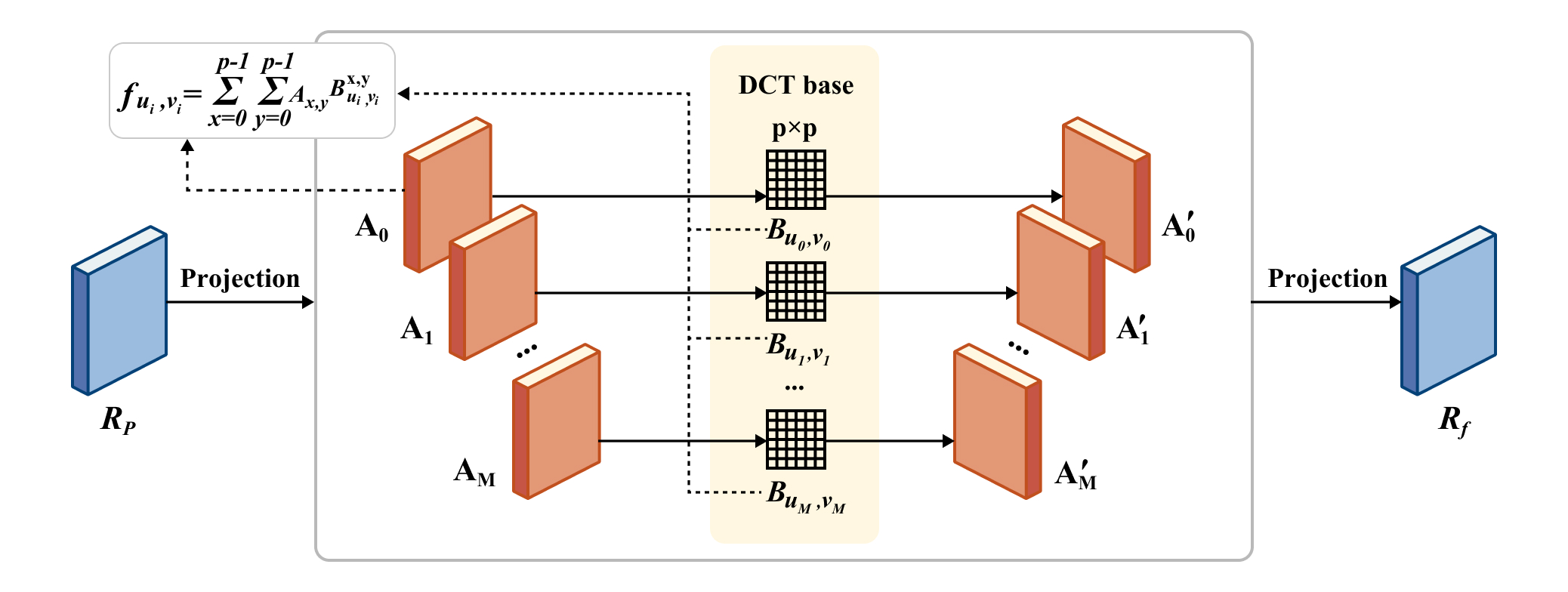}
	\caption{Structure of the multi-frequency mixer. The input feature map $R_p$ is split into $A_i$ along the channel dimension after projection mapping. $A_i$ is then transformed by the pre-selected DCT basis functions $B_{u_i,v_i} $ of the size $p \times p$ to obtain the frequency values $A'_i$. The output feature map $R_f$ is achieved by projection mapping after concatenating all $\{A'_i\}$ along the channel dimension.}
	\label{fig:mfm}
\end{figure*}
\subsection{Cross-Temporal Interactor}
\label{sec:method:cti}
Massive complex target objects and non-interest changes motivate us to emphasize changes of interest while suppressing non-interest changes in feature extraction. Previous methods extract bi-temporal features using Siamese neural networks, but lacking cross-temporal interaction causes the inefficacy of these features in identifying changes of interest. In fact, bi-temporal images as input can be regarded as a way to augment data, which emphasizes the feature differences in the changed areas by referencing the features in the other time. 

Recent work~\cite{asymmetric} implements the cross-attention mechanism in the change detection header to fuse bi-temporal features. However, high computation leads to limited satisfaction. Inspired by the gating mechanism, we propose to learn weights for enhancing the feature differences, so as to achieve the above-mentioned goal. Specifically, each CTI takes as input the bi-temporal features $F_1$ and $F_2$ corresponding to $T_1$ and $T_2$, and firstly adopts the element-level subtraction to obtain a coarse change representation $R_c$ as,
\begin{equation}
R _c = F_1 \ominus F_2,
\end{equation}
where $\ominus$ denotes the element-wise subtraction operation. Bi-temporal features $F_1$ and $F_2$ are then respectively concatenated with $R_c$, followed by the processing with the depth-wise separable convolution $\varphi(\cdot)$ and the Sigmoid activation function $\sigma(\cdot)$ to obtain the weight maps $W_1$ and $W_2$ as,
\begin{equation}
\begin{split}
&W _1 = \sigma(\varphi(F_1 \oplus R_c)),\\
&W _2 = \sigma(\varphi(F_2 \oplus R_c)),
\end{split}
\end{equation}
where $\oplus$ denotes the concatenation operation. This choice to use depthwise separable convolution (DSConv) is driven by two primary factors: (1) DSConv greatly reduces the model’s parameter count, thereby lowering computational cost during feature fusion, and (2) depthwise convolution in DSConv has been shown to enhance detail processing \cite{ren2022shunted}, allowing the model to capture finer spatial details and produce more accurate weighting during interaction. Finally, we adjust $F_1$ and $F_2$ with $W_1$ and $W_2$, respectively, to reach the difference-enhanced features $R_1$ and $R_2$ as,
\begin{equation}
\begin{split}
&R _1 = W _1 \times F _1,\\
&R _2 = W _2 \times F _2.
\end{split}
\end{equation}

\subsection{Cross-Spatial Interactor}
\label{sec:method:csi}
With the increasing network depth, mainstream classification backbones~\cite{resnet, vit, poolormer} often result in the loss of tiny objects and poorly distinguishable features, which hardly suit RSCD requiring high spatial detail. Despite being capable of weighing semantic information and spatial detail, the U-Net architecture has a fixed perceptual field and requires a high computational cost, whose application is limited. Hence, we devise the CSI, a per-stage feature extraction module based on the U-shaped architecture. Specifically, the CSI relies on base blocks for feature extraction. The base block follows the Transformer's structure, including a regularization function, a multi-frequency mixer and a channel MLP with residual connections. As mentioned, the deepest features of each CSI are the input to the corresponding CTI for feature separation using the gating mechanism, whereas the CTI enhances the bi-temporal features with temporal differences and returns them to the CSI. Afterwards, the multi-level feature fusion enables the features output from each stage to incorporate spatio-temporal interactions, making the feature more robust and discriminative. Finally, the MLP layer is designed to refine and enhance the extracted frequency domain features in a channel-wise manner, similar to the linear mapping layer used after multi-head attention. 

\setlength{\tabcolsep}{3.8pt}
\begin{table*}[t]
	\begin{center}
		\caption{
		Comparison of performance for RSCD on three datasets. Highest scores are in bold. All scores are in percentage.
		}
		\label{table:1}
            \begin{tabular}{l||ccccc||ccccc||ccccc}
		\Xhline{1.2pt}
            \rowcolor{mygray}
		     &\multicolumn{5}{c||}{WHU-CD} &\multicolumn{5}{c||}{LEVIR-CD} &\multicolumn{5}{c}{CLCD}\\
            \rowcolor{mygray}
			\multicolumn{1}{c||}{\multirow{-2}{*}{Method}}
                &F1 &Pre. &Rec. &IoU &OA  &F1 &Pre. &Rec. &IoU &OA &F1 &Pre. &Rec. &IoU &OA \\			
                \hline \hline
                FC-EF~\cite{fc-siam}  &72.01 &77.69 &67.10 &56.26 &92.07 &83.4 &86.91 &80.17 &71.53 &98.39 &48.64 &73.34 &36.29 &32.14 &94.30\\
   FC-Siam-Di~\cite{fc-siam}  &58.81 &47.33 &77.66 &41.66 &95.63 &86.31 &89.53 &83.31 &75.92 &98.67 &44.10 &72.97 &31.60 &28.29 &94.04\\
   FC-Siam-Conc~\cite{fc-siam} &66.63 &60.88 &73.58 &49.95 &97.04 &83.69 &91.99 &76.77&71.96&98.49&54.48&68.21&45.22&37.35&94.35\\
   IFNet~\cite{ifnet} &83.40 &\bf96.91 &73.19 &71.52 &98.83 &88.13 &\bf94.02 &82.93&78.77&98.87&48.65&49.96&47.41&32.14&92.55\\
   DTCDSCN~\cite{dtcdscn}&71.95 &63.92 &82.30 &56.19 &97.42 &87.67 &88.53&86.83&78.05&98.77&60.13&62.98&57.53&42.99&94.32 \\
   BIT~\cite{bit}&83.98 &86.64 &81.48 &72.39 &98.75 &89.31 &89.24 &89.37&80.68&98.92&57.13&64.39&51.34&39.99&94.27\\
   SNUNet~\cite{snunet}&83.50 &85.60 &81.49 &71.67 &98.71 &88.16 &89.18 &87.17&78.83&98.82&60.54&65.63&56.19&43.41&94.55\\
   ChangeStar(FarSeg)~\cite{changestar}&87.01 &88.78 &85.31 &77.00 &98.70 &89.30 &89.88 &88.72&80.66&98.90&60.75&62.23&59.34&43.63&94.30\\
   DMATNet~\cite{dmatnet}&85.07 &89.46 &82.24 &74.98 &95.83 &89.97 &90.78 &89.17&81.83&98.06&66.56&72.74&61.34&49.87&95.41\\
   LGPNet~\cite{lgpnet}&79.75 &89.68 &71.81 &66.33 &98.33 &89.37 &93.07 &85.95&80.78&99.00&55.42&62.98&49.49&38.33&94.10 \\
   ChangeFormer~\cite{changeformer}&81.82 &87.25 &77.03 &69.24 &94.80 &90.40 &92.05&88.80&82.48&99.04&58.44&65.00&53.07&41.28&94.38\\
   DMINet~\cite{dminet}&88.69 &93.84 &86.25 &79.68 &98.97 &90.71 &92.52&89.95&82.99&99.07&-&-&-&-&-\\
   USSFC-Net \cite{USSFCNet}&88.93 & 91.56 & 86.43 & \bf80.06 & \bf99.01 & 88.80 & 87.18 &\bf90.49 & 79.86 & 98.84&63.04 & 64.83 & 61.34 & 46.03 &  94.42 \\
   			\hline
			\textbf{STeInFormer} (Ours) &\bf89.61 &91.01 &\bf88.26 &79.87 &98.68 &\bf91.47 &93.37 &89.65 &\bf83.03 &\bf99.26 &\bf73.83 &\bf74.52 &\bf73.16 &\bf58.26 &\bf96.48\\
			\hline
		\end{tabular}
	\end{center}
\end{table*}
\setlength{\tabcolsep}{2pt}

\subsection{Multi-Frequency Mixer}
\label{sec:method:mfm}
Numerous non-interest changes, complex spatial objects and limited spectral information are the key RSCD issues yet to be addressed in spatial domain. Therefore, we attempt to explore frequency domain for RSCD, as the useful information on frequency is hardly observable in spatial domain. Meanwhile, conventional Transformer-based methods rely on  attention mechanisms~\cite{vit, swin} or MLPs~\cite{mlp, mlps} for token mixing, but their feasibility is limited by high time and space complexities. Inspired by~\cite{fcanet} and multi-head attention mixers, we design a novel multi-frequency mixer that combines prior frequency to achieve the information interaction in frequency domain and, therefore, linear complexity.

\setlength{\tabcolsep}{12pt}
\begin{table}[t]
	\begin{center}
		\caption{
		Comparison of space complexity and computational cost. Input images were resized to $256\times256\times3$. The best values are in bold.
		}
		\label{table:efficiency}
		\begin{tabular}{l||cc}
                \Xhline{1.2pt}
			\rowcolor{mygray}
			Method & Params (M) &Flops (G)\\
			\hline \hline
			 IFNet~\cite{ifnet} & 50.44 &82.26\\
			 DTCDSCN~\cite{dtcdscn} &41.07 &14.42\\
			 LGPNet~\cite{lgpnet} &70.99 &125.79 \\
			 DMATNet~\cite{dmatnet} &13.27 &- \\
			 SNUNet~\cite{snunet}&12.03 &54.88 \\
              BIT~\cite{bit} &3.55 &10.60\\
              ChangeFormer~\cite{changestar} &267.90 &129.27\\
              DMINet~\cite{dminet} &6.24 &14.42\\
			 \hline
			 \textbf{STeInFormer} (Ours) &\bf1.26 &\bf9.42 \\
			\hline
		\end{tabular}
	\end{center}
\end{table}

\paragraph{Discrete Cosine Transform (DCT)} 
To facilitate the following analysis, we provide the basis functions of the 2D DCT as,
\begin{equation}
    B_{h, w}^{x, y}=\alpha_{h}\alpha_{w}\cos\frac{\pi(2x+1)h}{2H}\cos\frac{\pi(2y+1)w}{2W},
\end{equation}
and
\begin{equation}
\alpha_{h}=\left\{\begin{array}{l l}{{1/\sqrt{H},}}&{{h=0}}\\ {{\sqrt{2/H},}}&{{1\leq h\leq H-1,}}\end{array}\right.
\end{equation}
\begin{equation}
    \alpha_{w}=\left\{\begin{array}{l l}{{1/\sqrt{W},}}&{{w=0}}\\ {{\sqrt{2/W},}}&{{1\leq w\leq W-1.}}\end{array}\right.
\end{equation}
where $h$ and $w$ denote the height and width dimensions of the spatial position of a pixel feature within the feature map, respectively. $x$ and $y$ denote the position of the 2D DCT frequency spectrum, i.e., $x \in \{0, 1,\dots, H-1\}, y \in \{0, 1,\dots, W-1\}$. 
Then, the 2D DCT can be formulated as,
\begin{equation}
    f_{x, y}=\sum_{h=0}^{H-1} \sum_{w=0}^{W-1} A_{h, w} B_{h, w}^{x, y},
\end{equation}
where $A\in \mathbb{R} ^{H\times W}$ denotes the input image, $f\in \mathbb{R} ^{H\times W}$ represents the 2D DCT frequency spectrum, $H$ and $W$ are the height and width of $A$, and the inverse 2D DCT can be formulated as,
\begin{equation}
    A_{h, w}=\sum_{x=0}^{H-1} \sum_{y=0}^{W-1} f_{x, y} B_{h, w}^{x, y}.
\end{equation}

\setlength{\tabcolsep}{3pt}
\begin{table}[t]
	\begin{center}
		\caption{
		Comparison of performance using our STeInFormer as the backbone of other methods or not. The highest scores are in bold. 
		}
		\label{table:block}
		\begin{tabular}{l|c||c||c||c}
                \Xhline{1.2pt}
			\rowcolor{mygray}
			Method &Backbone &WHU-CD &LEVIR-CD &CLCD\\
			\hline \hline
			 IFNet~\cite{ifnet} & VGG16~\cite{vgg} &83.40&88.13 &48.65\\
                    & STeInFormer &\bf89.81&\bf91.74&\bf74.05 \\
                \hline
                BIT~\cite{bit}& ResNet18~\cite{resnet} &83.98&89.31 &57.13  \\
                & STeInFormer &\bf90.21 &\bf92.07 &\bf74.32 \\
                \hline
                SNUNet~\cite{snunet}&NestedUNet~\cite{snunet}&83.50 &88.16 &60.54\\
                & STeInFormer &\bf90.16&\bf91.85 &\bf74.28 \\
                \hline
		\end{tabular}
	\end{center}
 \label{tab:backbone}
\end{table}

The frequency spectrum of an image can be obtained using the 2D DCT. Each frequency value can be regarded as a feature for a certain pattern of the image. For example, low frequencies express structural detail and high frequencies represent textural detail. We select multiple valid frequency values and pass them into the proposed method as the result of token mixing. 

\paragraph{Efficient batch processing} 
Considering the inefficiency of multiple DCTs in batch processing, we provide an efficient implementation as follows. As shown in Fig.~\ref{fig:mfm}, we select the bases $B_{u_i,v_i}$ of $M$ DCTs in advance. 
${u_i,v_i}$ denote the frequency component 2D indices that we have filtered from $x,y$. $i \in \{0, 1,\dots, M-1\}$, where M denotes the total frequency numbers we have selected. 
Specially, we first obtain $A_i$ from the input feature map $R_p$ through projection (i.e., $1\times 1$ convolution in our implementation) and splitting operations. Projection is implemented to increase the feature dimension for subsequent multi-frequency extraction operations. This is a similar operation of multi-head attention \cite{transformer}, and can enhance the expressive power of the model. Then, for each frequency head $i$, $A_i$ is weighted and summed with $B_{u_i,v_i}$ (it is implemented by $p \times p$ convolution) to calculate the corresponding frequency values $A'_i$. 
 Afterwards, all $\{A'_i\}$ are stitched and projected to produce a feature map $R_f$ as the mixer's output. The similarity between our implementation and group convolution~\cite{group} reflects its efficiency for batch processing.
\begin{figure*}[t]
\centering
\captionsetup[subfloat]{labelsep=none,format=plain,labelformat=empty}
\subfloat[$T_1$]{
\begin{minipage}[t]{0.095\linewidth}
\includegraphics[width=1\linewidth]{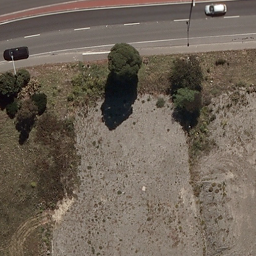}\vspace{1pt}
\includegraphics[width=1\linewidth]{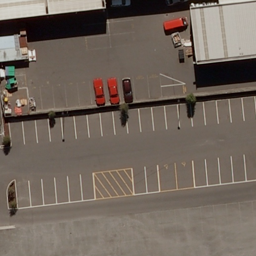}\vspace{3pt}
\includegraphics[width=1\linewidth]{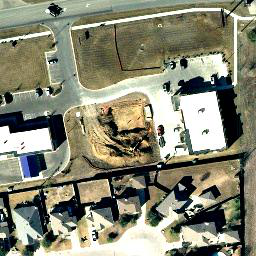}\vspace{1pt}
\includegraphics[width=1\linewidth]{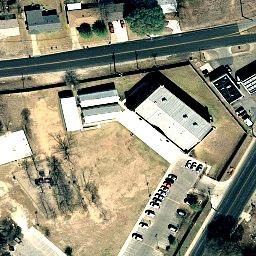}\vspace{3pt}
\includegraphics[width=1\linewidth]{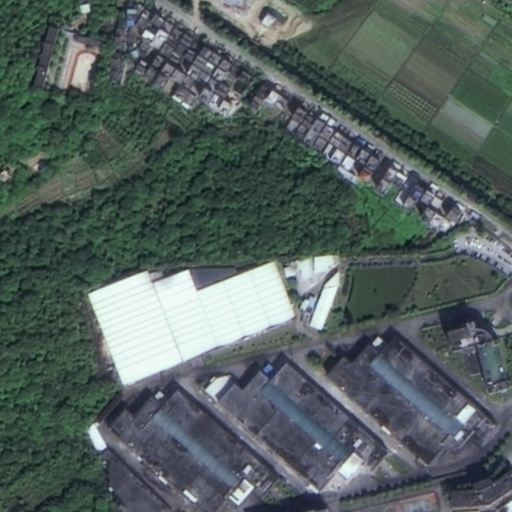}\vspace{1pt}
\includegraphics[width=1\linewidth]{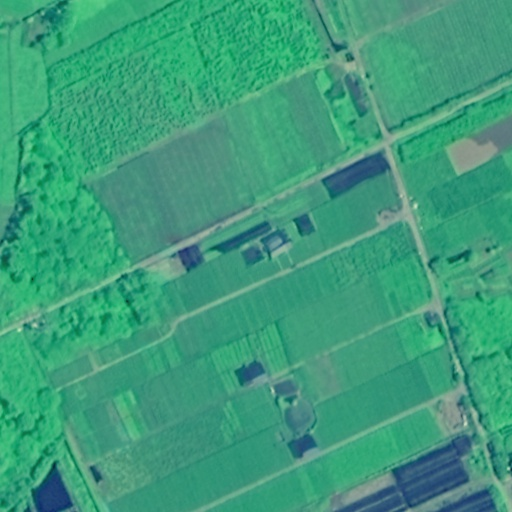}\vspace{3pt}
\end{minipage}}
\subfloat[$T_2$]{
\begin{minipage}[t]{0.095\linewidth}
\includegraphics[width=1\linewidth]{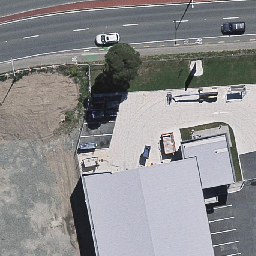}\vspace{1pt}
\includegraphics[width=1\linewidth]{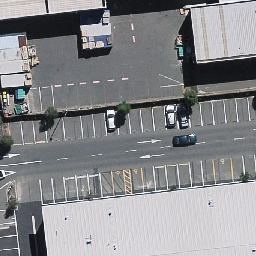}\vspace{3pt}
\includegraphics[width=1\linewidth]{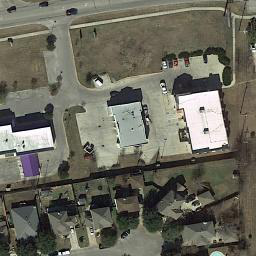}\vspace{1pt}
\includegraphics[width=1\linewidth]{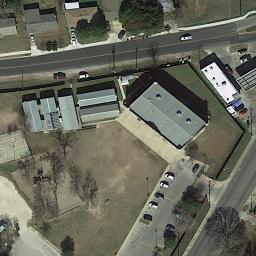}\vspace{3pt}
\includegraphics[width=1\linewidth]{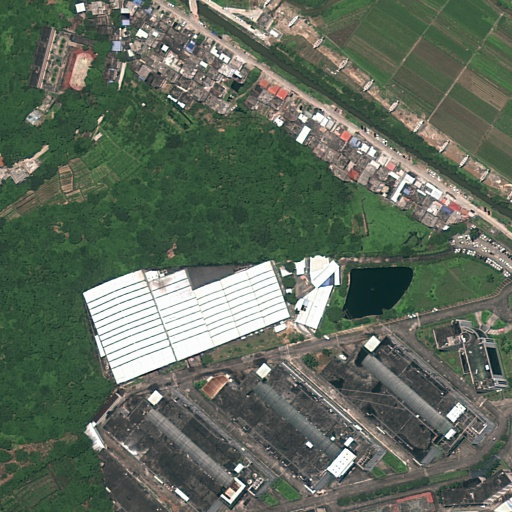}\vspace{1pt}
\includegraphics[width=1\linewidth]{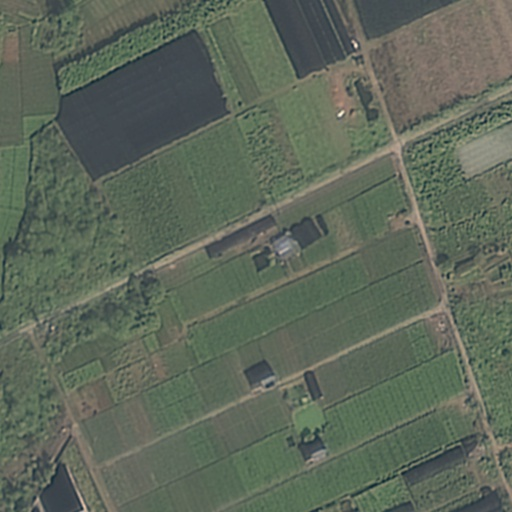}\vspace{3pt}
\end{minipage}}
\subfloat[GT]{
\begin{minipage}[t]{0.095\linewidth}
\includegraphics[width=1\linewidth]{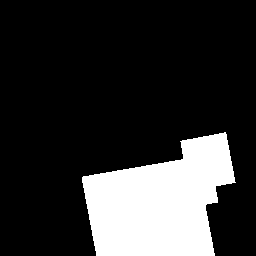}\vspace{1pt}
\includegraphics[width=1\linewidth]{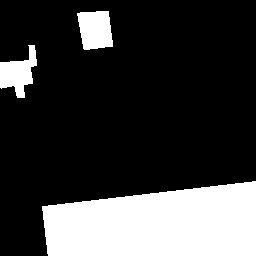}\vspace{3pt}
\includegraphics[width=1\linewidth]{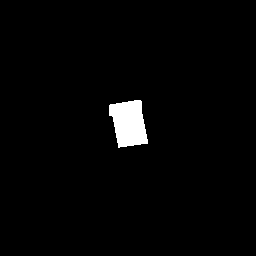}\vspace{1pt}
\includegraphics[width=1\linewidth]{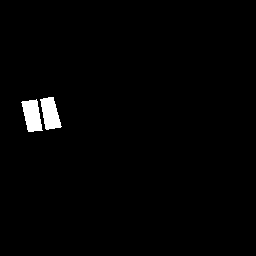}\vspace{3pt}
\includegraphics[width=1\linewidth]{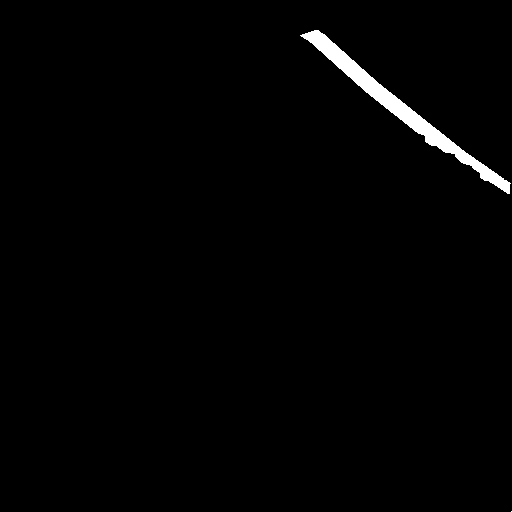}\vspace{1pt}
\includegraphics[width=1\linewidth]{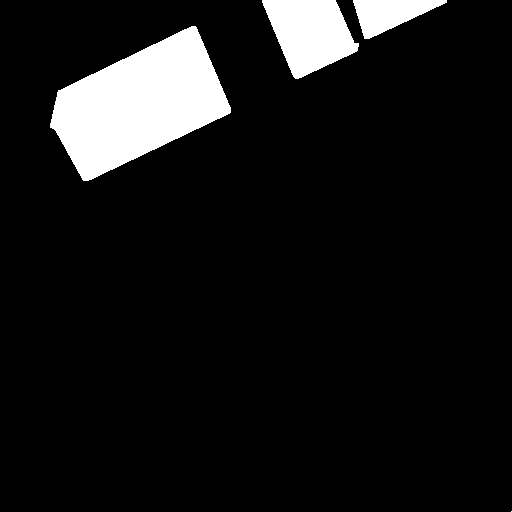}\vspace{3pt}
\end{minipage}}
\subfloat[FC-Siam-diff]{
\begin{minipage}[t]{0.095\linewidth}
\includegraphics[width=1\linewidth]{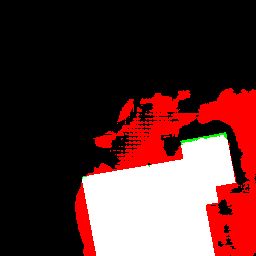}\vspace{1pt}
\includegraphics[width=1\linewidth]{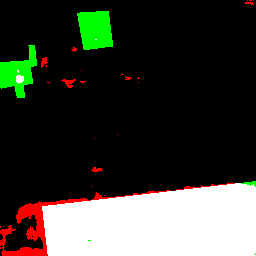}\vspace{3pt}
\includegraphics[width=1\linewidth]{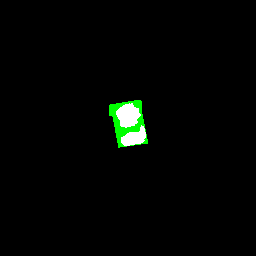}\vspace{1pt}
\includegraphics[width=1\linewidth]{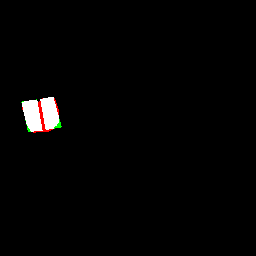}\vspace{3pt}
\includegraphics[width=1\linewidth]{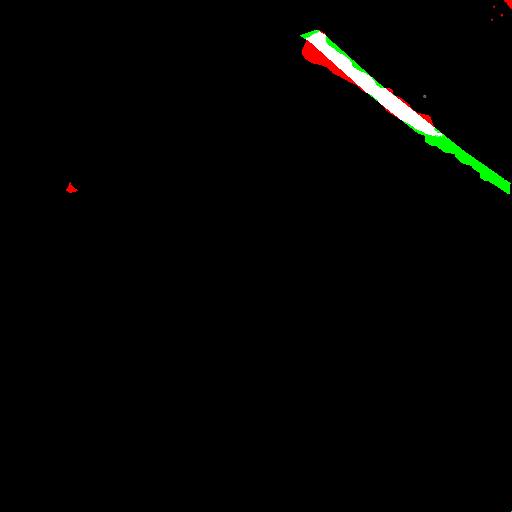}\vspace{1pt}
\includegraphics[width=1\linewidth]{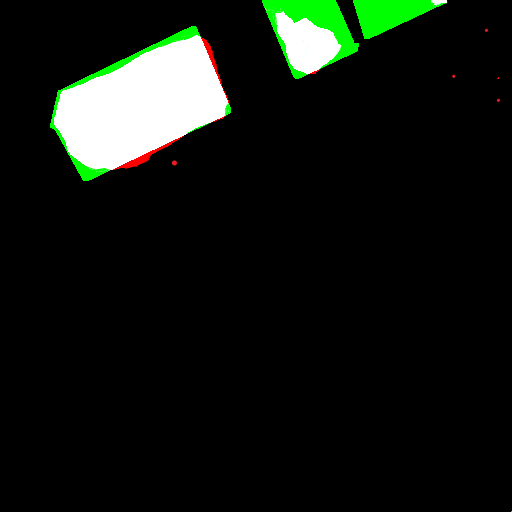}\vspace{3pt}
\end{minipage}}
\subfloat[FC-Siam-Conc]{
\begin{minipage}[t]{0.095\linewidth}
\includegraphics[width=1\linewidth]{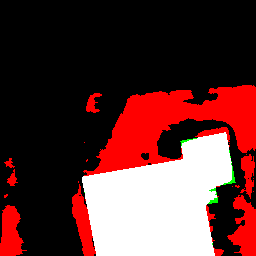}\vspace{1pt}
\includegraphics[width=1\linewidth]{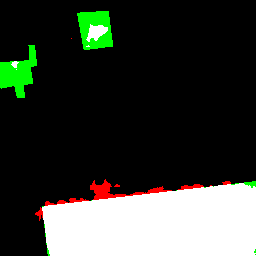}\vspace{3pt}
\includegraphics[width=1\linewidth]{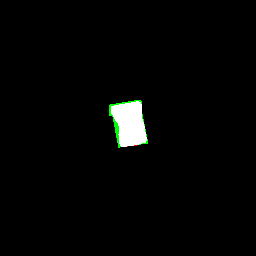}\vspace{1pt}
\includegraphics[width=1\linewidth]{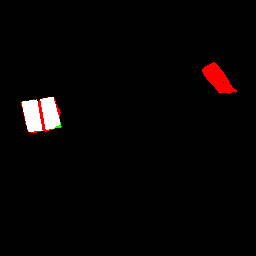}\vspace{3pt}
\includegraphics[width=1\linewidth]{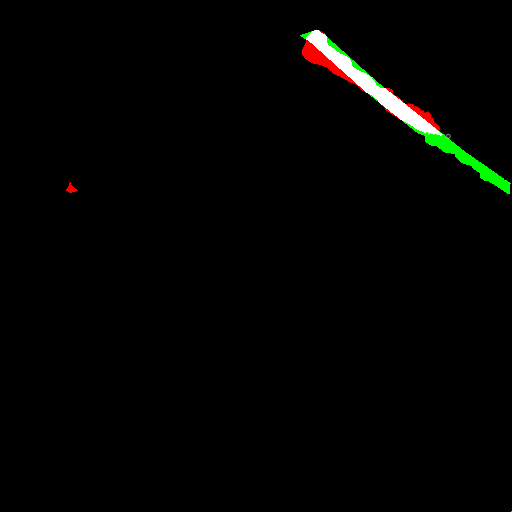}\vspace{1pt}
\includegraphics[width=1\linewidth]{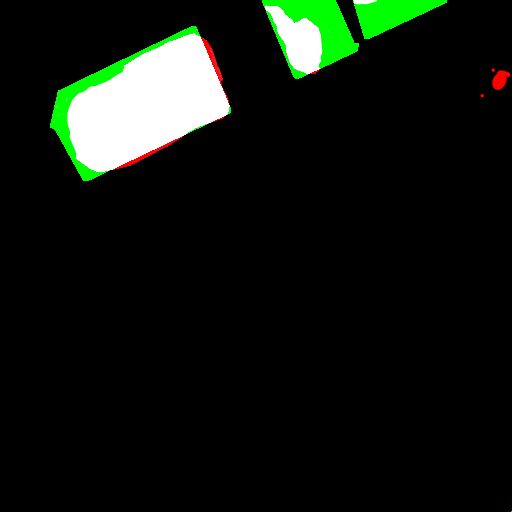}\vspace{3pt}
\end{minipage}}
\subfloat[SNUNet]{
\begin{minipage}[t]{0.095\linewidth}
\includegraphics[width=1\linewidth]{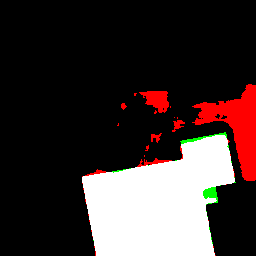}\vspace{1pt}
\includegraphics[width=1\linewidth]{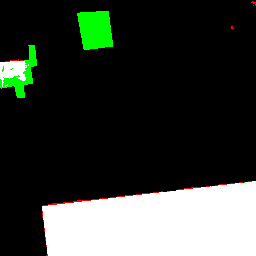}\vspace{3pt}
\includegraphics[width=1\linewidth]{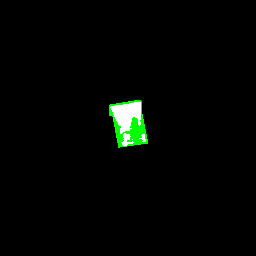}\vspace{1pt}
\includegraphics[width=1\linewidth]{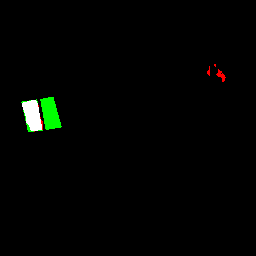}\vspace{3pt}
\includegraphics[width=1\linewidth]{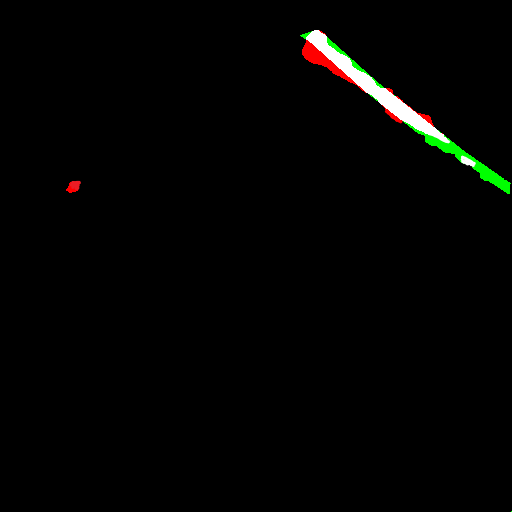}\vspace{1pt}
\includegraphics[width=1\linewidth]{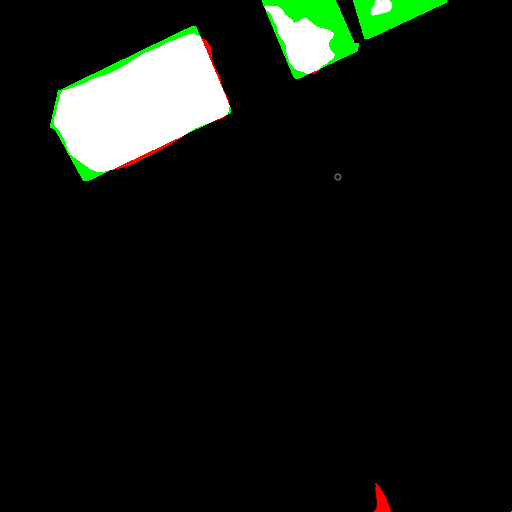}\vspace{3pt}
\end{minipage}}
\subfloat[BIT]{
\begin{minipage}[t]{0.095\linewidth}
\includegraphics[width=1\linewidth]{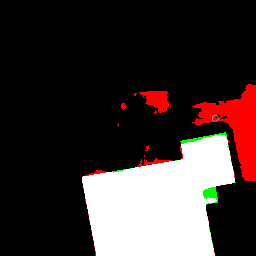}\vspace{1pt}
\includegraphics[width=1\linewidth]{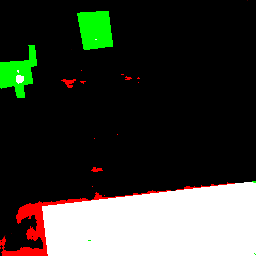}\vspace{3pt}
\includegraphics[width=1\linewidth]{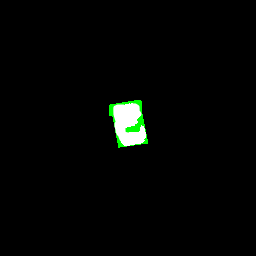}\vspace{1pt}
\includegraphics[width=1\linewidth]{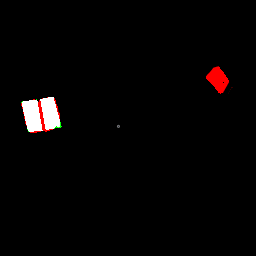}\vspace{3pt}
\includegraphics[width=1\linewidth]{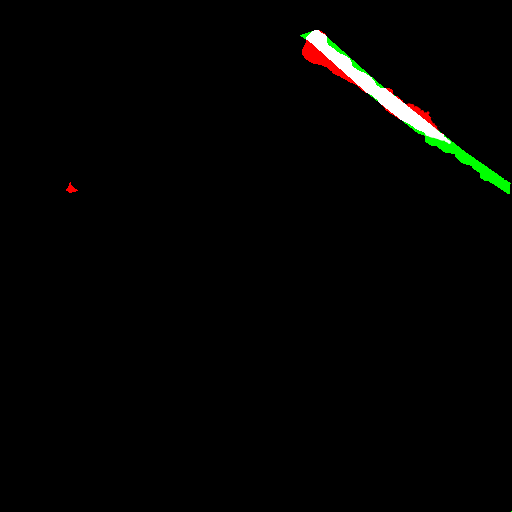}\vspace{1pt}
\includegraphics[width=1\linewidth]{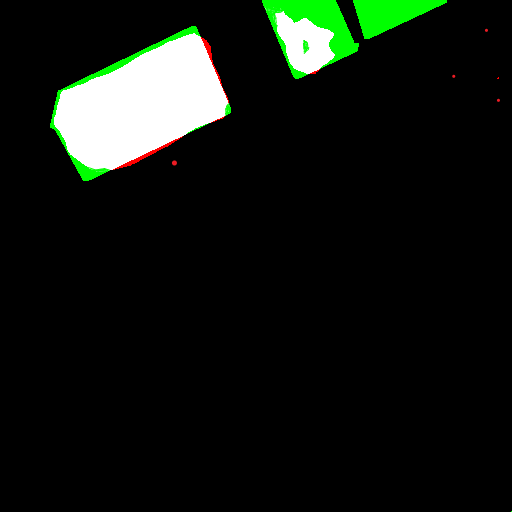}\vspace{3pt}
\end{minipage}}
\subfloat[LGPNet]{
\begin{minipage}[t]{0.095\linewidth}
\includegraphics[width=1\linewidth]{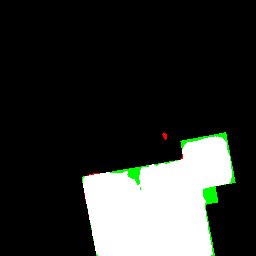}\vspace{1pt}
\includegraphics[width=1\linewidth]{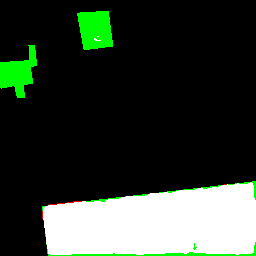}\vspace{3pt}
\includegraphics[width=1\linewidth]{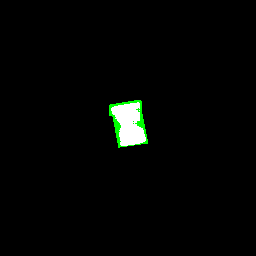}\vspace{1pt}
\includegraphics[width=1\linewidth]{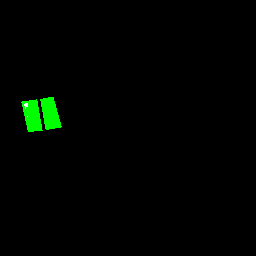}\vspace{3pt}
\includegraphics[width=1\linewidth]{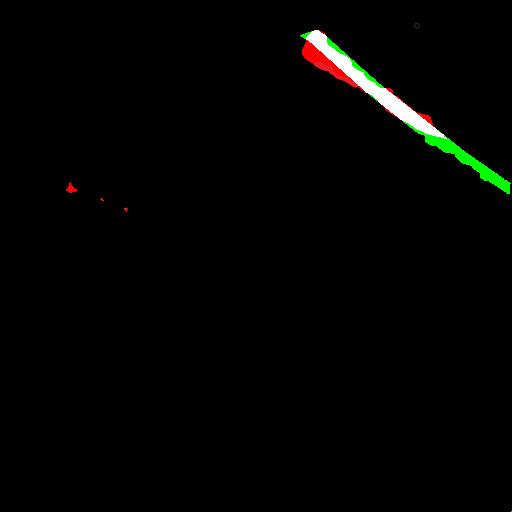}\vspace{1pt}
\includegraphics[width=1\linewidth]{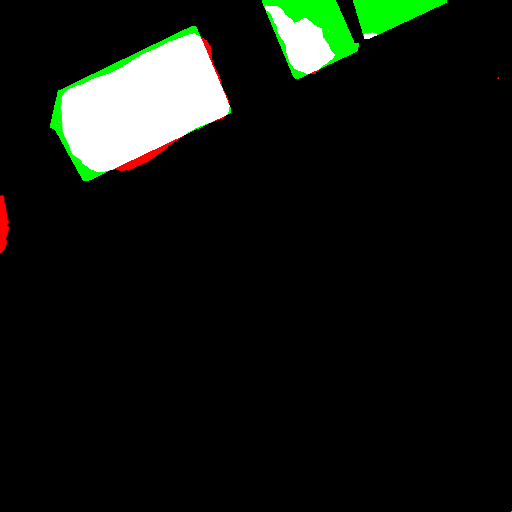}\vspace{3pt}
\end{minipage}}
\subfloat[ChangeFormer]{
\begin{minipage}[t]{0.095\linewidth}
\includegraphics[width=1\linewidth]{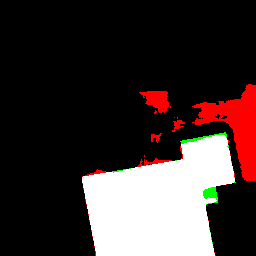}\vspace{1pt}
\includegraphics[width=1\linewidth]{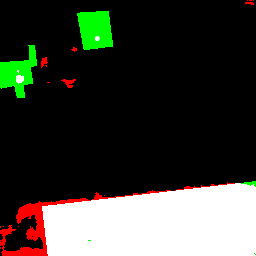}\vspace{3pt}
\includegraphics[width=1\linewidth]{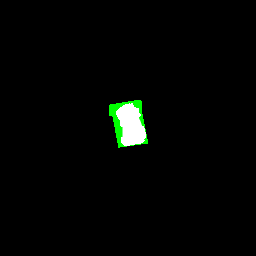}\vspace{1pt}
\includegraphics[width=1\linewidth]{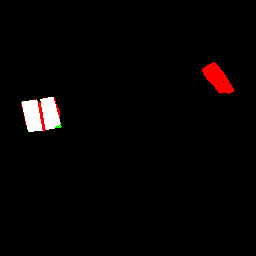}\vspace{3pt}
\includegraphics[width=1\linewidth]{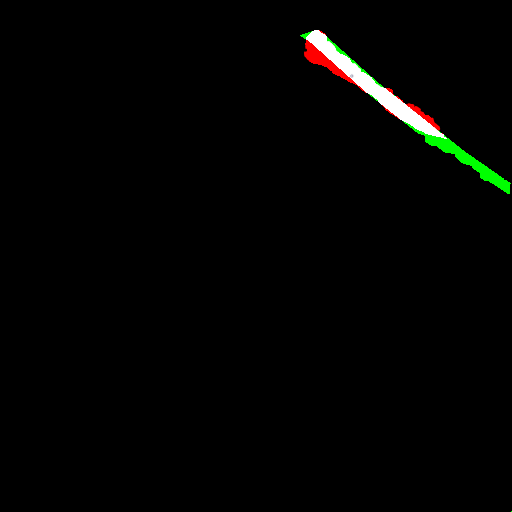}\vspace{1pt}
\includegraphics[width=1\linewidth]{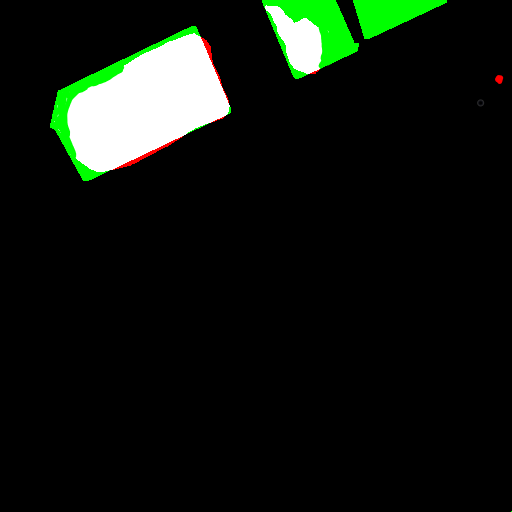}\vspace{3pt}
\end{minipage}}
\subfloat[STeInFormer]{
\begin{minipage}[t]{0.095\linewidth}
\includegraphics[width=1\linewidth]{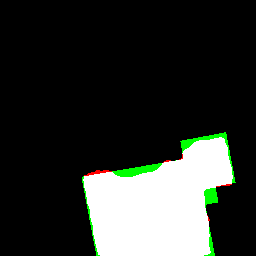}\vspace{1pt}
\includegraphics[width=1\linewidth]{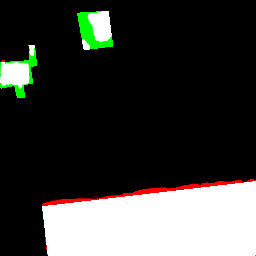}\vspace{3pt}
\includegraphics[width=1\linewidth]{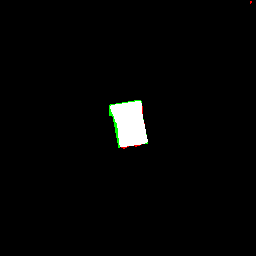}\vspace{1pt}
\includegraphics[width=1\linewidth]{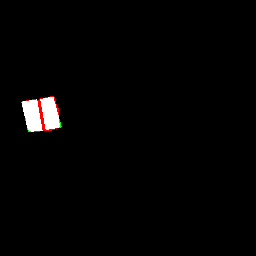}\vspace{3pt}
\includegraphics[width=1\linewidth]{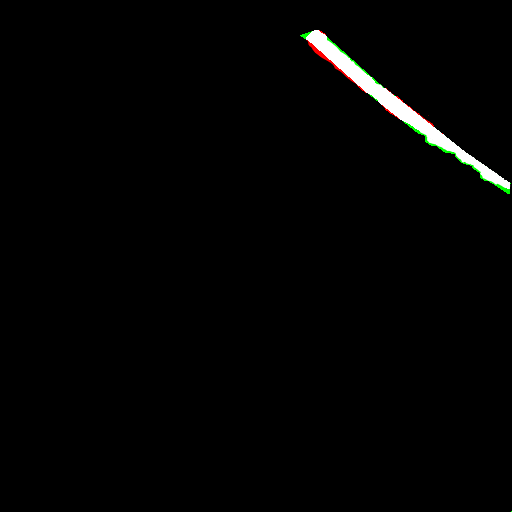}\vspace{1pt}
\includegraphics[width=1\linewidth]{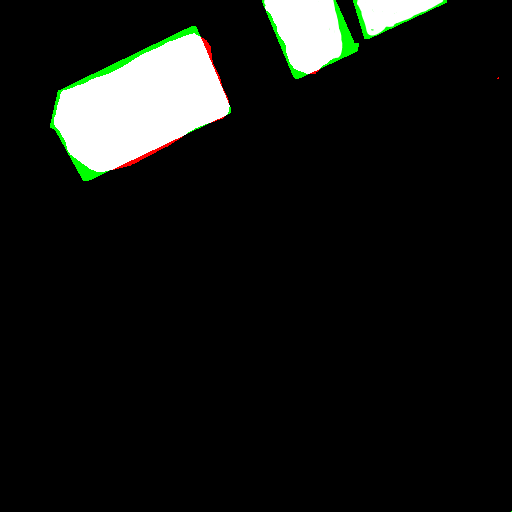}\vspace{3pt}
\end{minipage}}
\caption{Example outputs from our STeInFormer and other methods for comparison on WHU-CD (first and second rows), LEVIR-CD (third and fourth rows), and CLCD (fifth and sixth rows). Pixels are colored for visualization (white: true positive; black: true negative; red: false positive; green: false negative).}
\label{fig:res}
\end{figure*}

\paragraph{Frequency selection strategy} 
Apparently, the way of selecting frequency values impacts the performance of our STeInFormer for RSCD. We present the following strategies for frequency selection:
\begin{itemize}
    \item \textbf{Pre-trained priors strategy.} We conduct experiments on the ImageNet following~\cite{fcanet} to explore the importance of frequency by selecting only one frequency at a time. According to the experimental results, we select a number of the most important frequency values for token mixing.
    \item \textbf{Random selection strategy.} Since the signal energy tends to remain low-frequency, we randomly select several frequency values for token mixing while keeping the lowest ones.
    \item \textbf{Dynamic assignment strategy.} In contrast to a fixed frequency prior, we incorporate frequency selection into the training. Specifically, the spectral map is fed into a convolution module and obtain the weights with the Sigmoid activation function, from which we select the frequency values with the highest weights.
\end{itemize}

In our implementation of the STeInFormer, we adopt the pre-trained priors strategy. For the comparison of the above-discussed strategies, please refer to Sec.~\ref{sec:exp} for details.

\subsection{Decoder and Loss Function}
\label{sec:method:loss}

For the efficient design, our STeInFormer is able to achieve outstanding performance even with a simple MLP decoder. Specifically, the bi-temporal features output from each of four stages are concatenated into a change representation at the beginning. All four change representations are then upsampled into the same resolution (i.e., $H/2 \times W/2$) via bilinear interpolation followed by concatenation. Finally, the concatenated change representations are processed by $1\times 1$ convolution and 2-fold upsampling to generate the final change map.

Considering the significantly unbalanced distribution of changed and unchanged pixels, we employ a hybrid loss function combining a focal loss~\cite{focal} and a dice loss~\cite{dice} as,
\begin{equation}
\mathcal{L}  =  \lambda_{focal} \mathcal{L} _{\text{focal}} + \lambda_{dice} \mathcal{L}_{\text{dice}}.
\end{equation}
where $\lambda_{focal}$ and $\lambda_{dice}$ represent the weighting parameters. In our implementation, we use a ratio of 1:1 for these parameters.

The focal loss is formulated as,
\begin{equation}
\mathcal{L}_{\text{focal}} = -\alpha(1-\hat{p})^{\gamma}log(\hat{p}),
\end{equation}
\begin{equation}
\hat{p}=\left \{\begin{array}{ll}p,&\text{if\quad} y=1 \\
                            1-p,&\text{otherwise,} \\
                \end{array}
        \right.
\end{equation}
where $\alpha$ and $\gamma$ are two hyperparameters controlling the weights of positive and negative samples and the method's attention to difficult samples for detection, respectively, $p$ is the probability and $y$ is the pixel's binary label (0 or 1) corresponding to unchanged and changed. 

The dice loss is formulated as,
\begin{equation}
\mathcal{L}_{\text{dice}}  =1 - \frac{2 \cdot E \cdot softmax({E}')}{E+softmax({E}')},
\end{equation}
\begin{equation}
{E}' = \left\{{e'_{k}}\right\}, \ k\in \left[1,H\times W\right],
\end{equation}
where $E$ denotes the ground truth, ${E}'\in \mathbb{R} ^{H\times W \times 2}$ represents the change map, and $e'_{k}\in \mathbb{R} ^2$ refers to a pixel in ${E}'$.

\begin{figure*}[t]
	\centering
	\subfloat[WHU-CD]{
    \begin{minipage}[t]{0.32\linewidth}
    \includegraphics[width=1\linewidth]{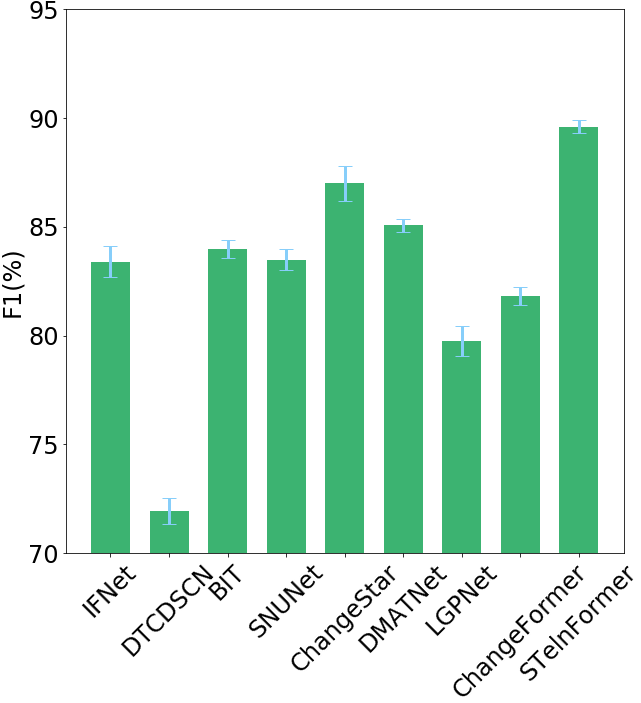}\vspace{4pt}
    \end{minipage}}
	\subfloat[LEVIR-CD]{
    \begin{minipage}[t]{0.32\linewidth}
    \includegraphics[width=1\linewidth]{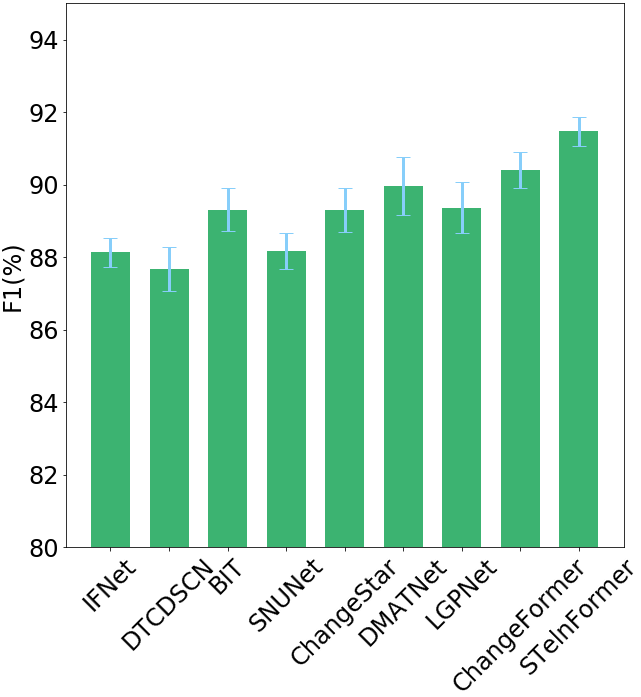}\vspace{4pt}
    \end{minipage}}
	\subfloat[CLCD]{
    \begin{minipage}[t]{0.32\linewidth}
    \includegraphics[width=1\linewidth]{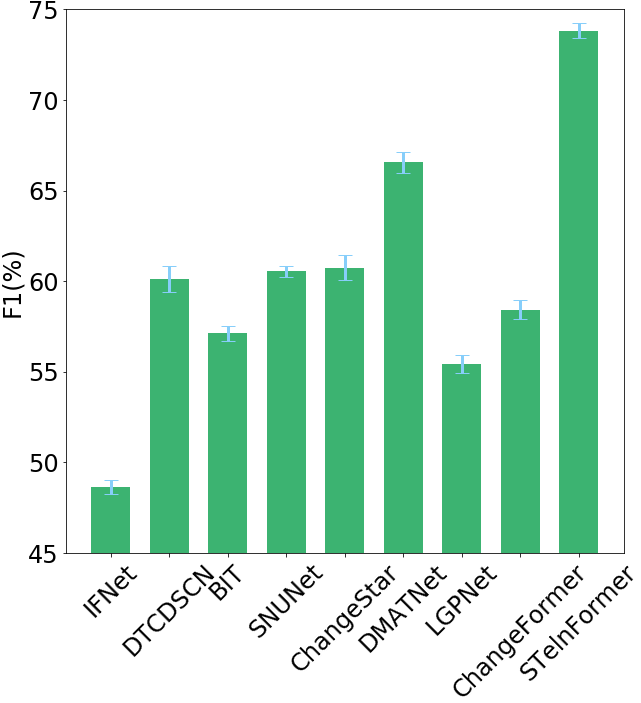}\vspace{4pt}
    \end{minipage}}
	\caption{Comparison of performance statistics on three datasets. Bars represent F1-scores, with standard deviations at the top. The bar corresponds to the middle of each method name.}
	\label{fig:zhu}
\end{figure*}

\section{Experiments}
\label{sec:exp}

We implement our STeInFormer using Python and PyTorch on a workstation with two NVIDIA GTX A6000 graphics cards (96~GB GPU memory in total). 
For the CSI module, we set the number of base blocks $N$ at different resolutions (i.e., 1/2, 1/4, 1/8, and 1/16 of the original image) to be 1. In addition, we set the channel number of the layer from higher to lower resolution to be 32, 48, 64, and 96, and therefore the number of channels for the CTI module is also set to be 96. In order to minimize the redundancy of the parameters, we set the expansion ratio of MLP to 2. 
The initial learning rate is $1e-4$ and the Adam~\cite{adam} optimizer is adopted with a weight decay of $1e-5$. During the training period, we employ a multi-step learning rate decay strategy with the gamma of $0.94$. The batch size is set to 4. We conduct data augmentation by flipping and rotating the images for training.

\subsection{Datasets and Metrics}
We evaluate our STeInFormer on three benchmark datasets in five common metrics, including F1-score (F1), recall (Rec), precision (Pre), intersection over union (IoU), and overall accuracy (OA).

WHU-CD~\cite{whu} is a public building CD dataset containing a pair of high-resolution (0.075m) aerial images of size $32507\times 15354$. For a corresponding data segmentation solution has not been provided, we crop the images into patches of size $256\times 256$ and randomly divide them into a training set ($60964$ images), a validation set ($762$ images), and a test set ($762$ images).

LEVIR-CD~\cite{levir} is a RSCD dataset containing 637 pairs of aligned remote sensing images of size $1024\times 1024$. These images have a time span of $5-14$ years and a spatial resolution of $0.5$~m. Following the default settings of the dataset, we crop the images into non-overlapping blocks of size $256\times 256$ and randomly divide them into a training set ($7120$ images), a validation set ($1024$ images) and a test set ($2048$ images).

The CropLand Change Detection (CLCD) dataset~\cite{clcd} contains $600$ pairs of remote sensing images of size $512\times 512$ and with a spatial resolution of $0.5-2$~m. We randomly divide the images into a training set ($360$ images), a validation set ($120$ images), and a test set ($120$ images).

\subsection{Comparison with State-of-the-Art Methods}
To validate the STeInFormer's effectiveness, we compare it with several state-of-the-art methods based on CNNs~\cite{fc-siam}, attention mechanisms~\cite{snunet,ifnet,dtcdscn,dmatnet}, Transformer~\cite{bit,changeformer}, transfer learning~\cite{lgpnet}, and reusing existent segmentation methods~\cite{changestar}. We implement the above networks using published code with default hyperparameter values. 

Experimental results on WHU-CD, LEVIR-CD and CLCD test sets are shown in Table~\ref{table:1}, suggesting that our STeInFormer significantly outperforms the other methods. For example, the proposed method achieves increases by $3.5/1.5/7.3$ and $6.8/1.1/15.4$ in F1-score compared to recently proposed DMATNet~\cite{dmatnet} and ChangeFormer~\cite{changeformer}, respectively, on three test sets. Besides, our STeInFormer obtaines higher F1-scores than the latest DMINet~\cite{dminet} (i.e., $0.9$ and $0.8$) on WHU-CD and LEVIR-CD test sets.

Fig.~\ref{fig:zhu} further illustrates the superiority of the proposed method to the other methods in F1-score as well as the standard deviation.
This can be explained by that incorporating the spatio-temporal interaction of bi-temporal features in feature extraction benefits the extraction of more robust and discriminative features, which enables our STeInFormer to be outstanding even with a simple MLP decoder. In addition, the improvement provided by the proposed method is particularly considerable on the CLCD dataset. We interpret it as spatio-temporal interaction can make more significant contributions to RSCD in the case of small data size and large spatial scale, which characterizes the use of CLCD dataset in our experiments.

\subsection{Efficiency comparisons}

We record the number of parameters in million (refered to as \emph{Params (M)}) and the number of floating-point operations per second measured in giga (refered to as \emph{Flops (G)}), so as to measure the complexity and computational cost of our STeInFormer and other methods for comparison. As shown in Table~\ref{table:efficiency}, the proposed method requires significantly less parameters and the least computational cost, while outperforming the other methods for RSCD. 
For example, compared to the SOTA method DMINet, STeINFormer requires only 20\% of the parameters and 65\% of the FLOPs, but has an average F1 improvement of 4 on the three change detection datasets. Even compared to the lightweight model BIT, STeINFormer still has better parameter efficiency. We attribute the parameter efficiency to the following three aspects: 1), we input only the lowest resolution bi-temporal image features (i.e., 1/32 of the original images) into the CTI module; 2), we use pre-trained frequencies as token mixers, which do not introduce additional learnable parameters and thus significantly reduce the number of parameters of the model; and 3), some of our lightweight designs, including light convolutional attention and simple mlp decoder.

\setlength{\tabcolsep}{4.5pt}
\begin{table*}[t]
	\begin{center}
		\caption{
		Comparison of performance by frequency selection strategy, method design, and number of frequency values.
		}
		\label{table:ablation}
            \begin{tabular}{cc||ccccc||ccccc||ccccc}
		\Xhline{1.2pt}
            \rowcolor{mygray}
		    & &\multicolumn{5}{c||}{WHU-CD} &\multicolumn{5}{c||}{LEVIR-CD} &\multicolumn{5}{c}{CLCD}\\
            \rowcolor{mygray}
			\multicolumn{1}{c}{\multirow{-2}{*}{CSI}}&\multicolumn{1}{c||}{\multirow{-2}{*}{CTI}}
                &F1 &Pre. &Rec. &IoU &OA  &F1 &Pre. &Rec. &IoU &OA &F1 &Pre. &Rec. &IoU &OA \\			
                \hline
                      \ding{51} & PP  &89.61 &\bf91.01 &88.26 &\bf79.87 &\bf98.68 &\bf91.47 &\bf93.37 &\bf89.65 &\bf83.03 &\bf99.26 &\bf73.83 &\bf74.52 &\bf73.16 &\bf58.26 &\bf96.48\\
   \ding{51} & RS &86.55&86.88 &86.23 &76.34 &96.33 &90.24 &91.24 &89.27 &81.44 &98.72&70.82 &71.27 &70.38 &54.65 &95.23\\
   \ding{51} & DA &\bf89.64 &90.87 &\bf88.44 &79.82 &98.57 &90.93 &92.48 &89.44 &82.85 &99.21 &71.32 &71.37 &71.28 &56.78 &94.84\\
   \hline
                \ding{55} & \ding{55}  &80.91& 81.34 &80.48 &72.33 &92.14   &87.15 &88.27 &86.06 &79.23 &98.28 &62.59&62.43&62.76&48.34 &94.97\\
   \ding{51} & \ding{55}  &86.89 &86.51 &87.26 &77.67 &97.11	   &89.86&92.18 &87.65 &81.58 &98.75 &69.29&70.76&67.88&54.67&95.52\\
   \ding{55} & \ding{51} &85.04 & 85.47 &84.62 &76.42 &95.37 &90.01 &91.97 &88.14 &81.47 &98.66 &69.36&69.31 &69.42&55.33 &95.68\\
   \ding{51} & \ding{51}&\bf89.61 &\bf91.01 &\bf88.26 &\bf79.87 &\bf98.68 &\bf91.47 &\bf93.37 &\bf89.65 &\bf83.03 &\bf99.26 &\bf73.83 &\bf74.52 &\bf73.16 &\bf58.26 &\bf96.48\\
			\hline
   Conv &\ding{51} &85.78 &84.34 &87.28 &75.21 &95.22&89.83 &90.23&89.44&81.11&98.88&69.27 &68.28 &70.29&52.23 &95.11\\
   M=1  &\ding{51} & 86.75&86.37 &87.14&76.46&96.53  &90.14&91.05 &89.24 &82.03 &99.11  &70.98& 71.63&70.34&54.76&95.31\\
    M=4 &\ding{51} &88.79 &89.43 &88.15 &79.23 &98.64  &90.87 &92.31 &89.48 &82.78 &99.14  &73.06 &73.28 &72.85 &57.01 &95.89\\
    M=8 &\ding{51}&\bf89.61 &\bf91.01 &\bf88.26 &\bf79.87 &\bf98.68 &91.47 &93.37 &\bf89.65 &\bf83.03 &\bf99.26 &\bf73.83 &\bf74.52 &\bf73.16 &\bf58.26 &\bf96.48\\
    M=16&\ding{51} & 88.92 &89.43 &88.15 &79.64 &98.63  &\bf91.52 &\bf93.48 &89.64 &82.83 &99.21  &73.31 &73.48 &73.15 &57.41 &96.06\\
			\hline
		\end{tabular}
	\end{center}
\end{table*}
\setlength{\tabcolsep}{2pt}

\subsection{Backbone comparisons}
Furthermore, we combine the STeInFormer with three existent methods to validate its capability to serve as a generic backbone for change detection tasks. 
Specifically, we first replace the CNN backbone of BIT (i.e., ResNet18), the deep feature extraction network of IFNet (i.e., VGG16), and the backbone of SNUNet (i.e., NestedUNet) with STeINFormer, respectively. Then, we keep the other components (e.g., the Bitemporal Image Transformer in BIT \cite{bit}, difference discrimination network in IFNet \cite{ifnet}, Skip-connection in SNUNet \cite{snunet}) unchanged.
As shown in Table~\ref{table:block}, these methods adopting the STeInFormer as the backbone achieve significant improvements on the three datasets, compared to their original configurations. However, these improvements are inconsiderable compared to our implementation of the STeInFormer (i.e., adopting a simple MLP decoder). We interpret this observation as our STeInFormer accomplish spatio-temporal interactions in feature extraction, while similar operations are also conducted by the decoders of the three methods, causing redundancy of interaction that may limit performance improvement for change detection tasks.

\setlength{\tabcolsep}{12pt}
\begin{table*}[t]
	\begin{center}
		\caption{
		Ablation of the Cross-Temporal Interactor design on the WHU-CD dataset. The best values are in bold. 
		}
		\label{table:attention}
		\begin{tabular}{l||cc|ccccc}
                \Xhline{1.2pt}
			\rowcolor{mygray}
			Variant & Params (M) &Flops (G) &F1-Score &Precision &Recall &IoU &OA\\
			\hline \hline
			 Attention&1.88&10.49&89.48&90.68&\bf88.31&79.34&98.41\\
			 Ours&\bf1.26&\bf9.42&\bf89.61&\bf91.01&88.26&\bf79.87&\bf98.68\\
			\hline
		\end{tabular}
	\end{center}
\end{table*}

\subsection{Qualitative visualization}
Fig. \ref{fig:res} provides the visualization of example RSCD results on the three datasets. It can be observed that our STeInFormer effectively detects the changes of interest (e.g., the images in the second and sixth rows, where unapparent building changes are also perceived) and to some extent refrains from being impacted by the non-interest changes (e.g., the images in the first and second rows, where the red areas are significantly reduced). Besides, our STeInFormer restores high-quality spatial structures and textural details of the changed objects. These findings support the effectiveness of incorporating the spatio-temporal interaction in feature extraction.

\begin{figure}[t]
	\centering
        \includegraphics[width=0.47\textwidth]{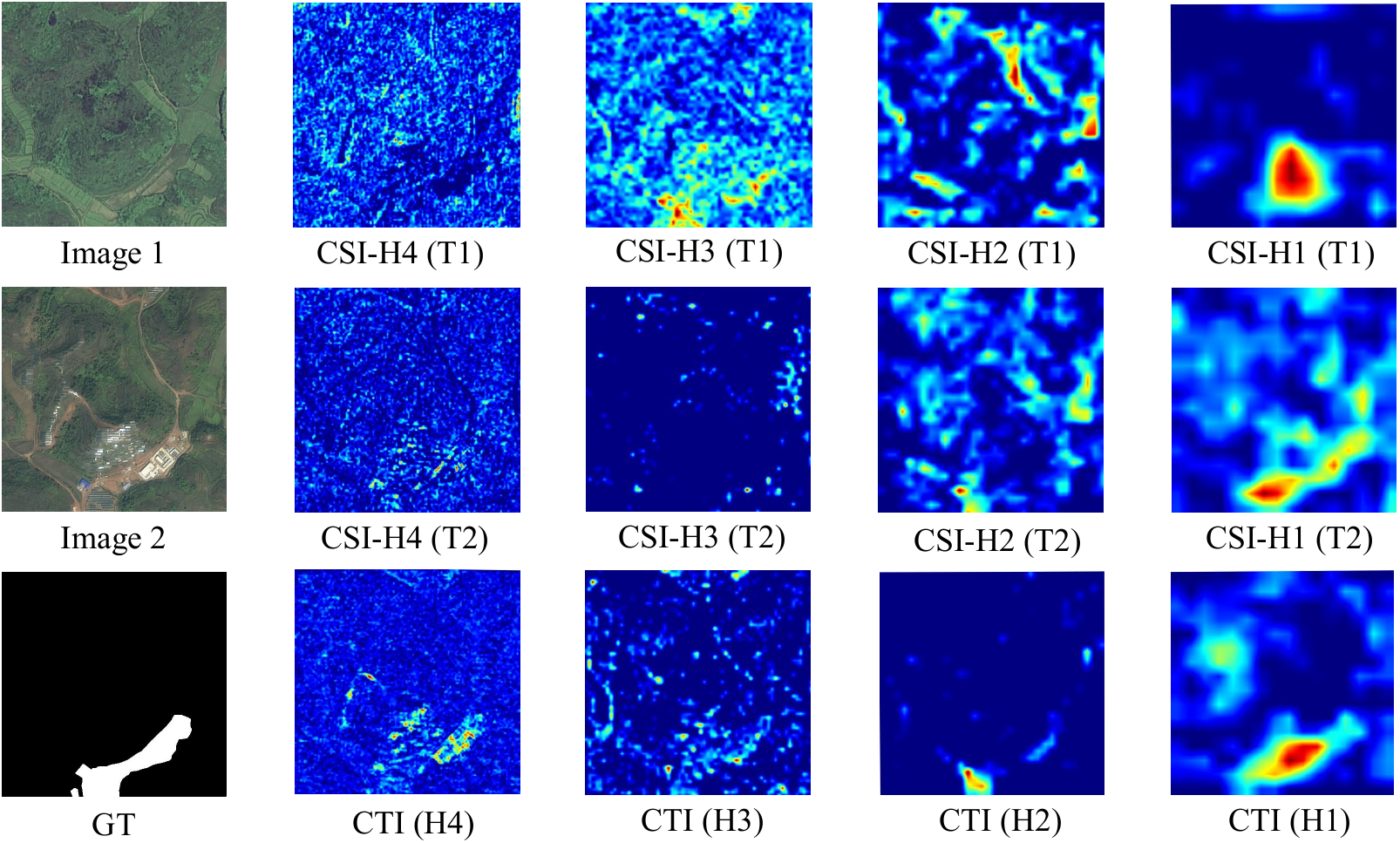}
	\centering
	\caption{Class activation maps for features of Cross-Spatial Interactor (CSI) and Cross-Temporal Interactor (CTI) in different layers. Example images are from the CLCD test set. H4, H3, H2, and H1 correspond to 1/2, 1/4, 1/8, and 1/16 resolution of the input image, respectively.}
\label{fig:cam_clcd}
\end{figure}

To explore the validity of the model, we additionally visualize the activation maps of the features output by the Cross-Temporal Interactor on the CLCD dataset, which is implemented based on Grad-CAM. As shown in Fig. \ref{fig:cam_clcd}, the activation values of the Cross-Temporal Interactor in the change region gradually increase as the depth of the model deepens, which proves that the model is effective in enhancing the semantic feature differences in the change region. 

\subsection{Ablation Studies}
We conduct a series of ablation experiments to determine the contribution of individual modules in the proposed method as well as the impact of parameter selection. As shown in Table~\ref{table:ablation}, we summarize the following findings.

\paragraph{Frequency selection strategy} We validate the three frequency selection strategies mentioned in Sec.~\ref{sec:method:mfm}, and the pre-trained priors (PP) strategy reaches the best results on the three datasets, whereas the random selection (RS) strategy corresponds to the worst. Besides, the dynamic assignment (DA) strategy is slightly inferior to the pre-trained priors strategy, which can be interpreted as that the dynamic assignment strategy provides insignificantly worse generalization due to training on limited data. Therefore, we select the frequency prior strategy for our STeInFormer to achieve better performance and linear complexity with prior frequency values.

\paragraph{Method design} We devise some variants to validate the efficacy of CSI and CTI. The base for the experiment is a Siamese neural network with tandem feature extraction modules at each stage without cross-temporal interaction, connected to the same MLP decoder adopted by our STeInFormer. Introducing CSI reorganizes the feature extraction modules at each stage into the U-shaped architecture. Introducing CTI adds the cross-temporal interaction to the feature extraction modules at each stage. The experimental results show adding either CSI or CTI enables a considerable improvement, and combining both further enlarges such an increase in performance.

\paragraph{Number of frequency values for the multi-frequency mixer.} It is clear that an overall stable outperformance could be reached on the three datasets when the number of frequency values (i.e., M) is $8$, although $16$ frequency values causes slighly higher F1-score and precision on the LEVIR-CD dataset only. Hence, we select $8$ as the number of frequency values for the multi-frequency mixer of our STeInFormer. 

\paragraph{Ablation of the Multi-Frequency Mixer.} We further explore the performance of using plain convolution (i.e., 3x3) instead of the multi-frequency mixer, which is shown in the Conv row in Table \ref{table:ablation}. The results show that when using only convolution, the model achieves F1 values of 85.78, 89.83 and 69.27 on WHU-CD, LEVIR-CD and CLCD datasets, respectively. Although we note that this result is very close to the performance when the frequency number is 1. However, when we increase the frequency number to 8, the expressive power of the proposed model is enhanced and significantly outperforms convolutional token mixer. This validates the effectiveness of the multi-frequency mixer.

\paragraph{Ablation of the Cross-Temporal Interactor.}
We further apply cross-attention to implement the cross-temporal interactor to validate the effectiveness of our convolutional gating mechanism. As shown in the Table \ref{table:attention}, the Attention variant is able to achieve an F1 value of 89.48, which is slightly lower than the convolutional gating mechanism version. However, Attention significantly increases the number of parameters (+0.62M) and the FLOPs (+1.07G) of the model, which is not conducive to a better balance between performance and efficiency.

\paragraph{Ablation of the weighting parameters in the loss function.}
To assess the impact of the weighting parameters on the model's performance, we have conducted additional experiments with different ratios of $\lambda_{focal}$ and $\lambda_{dice}$ on the WHU-CD dataset, as shown in Table \ref{table:lossf}. However, these configurations yield inferior performance compared to our chosen setup. Therefore, we ultimately select a 1:1 ratio.
\setlength{\tabcolsep}{3pt}
\begin{table}[t]
	\begin{center}
		\caption{
		Ablation of the weighting parameters on the WHU-CD dataset. The best values are in bold. 
		}
		\label{table:lossf}
		\begin{tabular}{cc||ccccc}
                \Xhline{1.2pt}
			\rowcolor{mygray}
			$\lambda_{focal}$ & $\lambda_{dice}$ &F1 &Pre. &Rec. &IoU &OA\\
			\hline \hline
			 1& 5 & 89.30 &  90.39  &  88.23  &  80.67  & 99.03\\
             1& 2 &89.53  &  92.44 &  86.79  &  \bf81.04  & \bf99.07\\ 
			 1 & 1 &\bf89.61&\bf91.01&88.26&79.87&98.68\\
             2& 1 &  89.16  &  87.99  &  \bf90.36  &  80.44  & 98.99 \\
             5& 1 & 88.40 & 87.96   &  88.85  &  79.21  & 98.93\\
			\hline
		\end{tabular}
	\end{center}
\end{table}

\section{Conclusion}
In this paper, we propose STeInFormer, a spatial-temporal interaction Transformer architecture to address two challenges in RSCD (i.e., frequent non-interest changes and the requirement for high spatial detail). Different from the existent RSCD methods, we abandon the paradigm of relying on a non-interactive Siamese neural network for feature extraction, but attempt to incorporate the spatio-temporal interactions of bi-temporal features in feature extraction. In addition, we design a parameter-free multi-frequency token mixer for token mixing, which for the first time tackles RSCD from the perspective of frequency domain. By combining frequency-valued priors, our multi-frequency token mixer is able to extract implicit frequency-domain features with linear time complexity. Experimental results on three datasets prove that our STeInFormer could significantly outperform other methods, while achieving a more satisfactory efficiency-accuracy trade-off regarding the number of parameters and the computational cost. In the future, we are to investigate a change detection head corresponding to our spatio-temporal interaction encoder architecture, which is expected to extract more useful information and knowledge for change detection tasks.

\bibliographystyle{IEEEtran}
\bibliography{refer}

\begin{thebibliography}{10}
\providecommand{\url}[1]{#1}
\csname url@samestyle\endcsname
\providecommand{\newblock}{\relax}
\providecommand{\bibinfo}[2]{#2}
\providecommand{\BIBentrySTDinterwordspacing}{\spaceskip=0pt\relax}
\providecommand{\BIBentryALTinterwordstretchfactor}{4}
\providecommand{\BIBentryALTinterwordspacing}{\spaceskip=\fontdimen2\font plus
\BIBentryALTinterwordstretchfactor\fontdimen3\font minus \fontdimen4\font\relax}
\providecommand{\BIBforeignlanguage}[2]{{%
\expandafter\ifx\csname l@#1\endcsname\relax
\typeout{** WARNING: IEEEtran.bst: No hyphenation pattern has been}%
\typeout{** loaded for the language `#1'. Using the pattern for}%
\typeout{** the default language instead.}%
\else
\language=\csname l@#1\endcsname
\fi
#2}}
\providecommand{\BIBdecl}{\relax}
\BIBdecl

\bibitem{cui2022unsupervised}
K.~Cui, R.~Li, S.~L. Polk, J.~M. Murphy, R.~J. Plemmons, and R.~H. Chan, ``Unsupervised spatial-spectral hyperspectral image reconstruction and clustering with diffusion geometry,'' in \emph{2022 12th Workshop on Hyperspectral Imaging and Signal Processing: Evolution in Remote Sensing (WHISPERS)}.\hskip 1em plus 0.5em minus 0.4em\relax IEEE, 2022, pp. 1--5.

\bibitem{cui2024real}
K.~Cui, W.~Tang, R.~Zhu, M.~Wang, G.~D. Larsen, V.~P. Pauca, S.~Alqahtani, F.~Yang, D.~Segurado, P.~Fine \emph{et~al.}, ``Real-time localization and bimodal point pattern analysis of palms using uav imagery,'' \emph{arXiv preprint arXiv:2410.11124}, 2024.

\bibitem{cui2024superpixel}
K.~Cui, R.~Li, S.~L. Polk, Y.~Lin, H.~Zhang, J.~M. Murphy, R.~J. Plemmons, and R.~H. Chan, ``Superpixel-based and spatially-regularized diffusion learning for unsupervised hyperspectral image clustering,'' \emph{IEEE Transactions on Geoscience and Remote Sensing}, 2024.

\bibitem{rstask}
L.~Khelifi and M.~Mignotte, ``Deep learning for change detection in remote sensing images: Comprehensive review and meta-analysis,'' \emph{Ieee Access}, vol.~8, pp. 126\,385--126\,400, 2020.

\bibitem{CMSCGC}
R.~Guan, Z.~Li, W.~Tu, J.~Wang, Y.~Liu, X.~Li, C.~Tang, and R.~Feng, ``Contrastive multiview subspace clustering of hyperspectral images based on graph convolutional networks,'' \emph{IEEE Transactions on Geoscience and Remote Sensing}, vol.~62, pp. 1--14, 2024.

\bibitem{SSGCC}
R.~Guan, W.~Tu, Z.~Li, H.~Yu, D.~Hu, Y.~Chen, C.~Tang, Q.~Yuan, and X.~Liu, ``Spatial-spectral graph contrastive clustering with hard sample mining for hyperspectral images,'' \emph{IEEE Transactions on Geoscience and Remote Sensing}, pp. 1--16, 2024.

\bibitem{monitor}
C.-F. Chen, N.-T. Son, N.-B. Chang, C.-R. Chen, L.-Y. Chang, M.~Valdez, G.~Centeno, C.~A. Thompson, and J.~L. Aceituno, ``Multi-decadal mangrove forest change detection and prediction in honduras, central america, with landsat imagery and a markov chain model,'' \emph{Remote Sensing}, vol.~5, no.~12, pp. 6408--6426, 2013.

\bibitem{urban}
C.~Marin, F.~Bovolo, and L.~Bruzzone, ``Building change detection in multitemporal very high resolution sar images,'' \emph{IEEE transactions on geoscience and remote sensing}, vol.~53, no.~5, pp. 2664--2682, 2014.

\bibitem{disaster}
S.~Mahdavi, B.~Salehi, W.~Huang, M.~Amani, and B.~Brisco, ``A polsar change detection index based on neighborhood information for flood mapping,'' \emph{Remote Sensing}, vol.~11, no.~16, p. 1854, 2019.

\bibitem{land}
X.~Li, F.~Ling, G.~M. Foody, and Y.~Du, ``A superresolution land-cover change detection method using remotely sensed images with different spatial resolutions,'' \emph{IEEE Transactions on Geoscience and Remote Sensing}, vol.~54, no.~7, pp. 3822--3841, 2016.

\bibitem{land2}
Y.~Hu, Y.~Dong \emph{et~al.}, ``An automatic approach for land-change detection and land updates based on integrated ndvi timing analysis and the cvaps method with gee support,'' \emph{ISPRS journal of photogrammetry and remote sensing}, vol. 146, pp. 347--359, 2018.

\bibitem{crossmatch}
R.~Liu, T.~Luo, S.~Huang, Y.~Wu, Z.~Jiang, and H.~Zhang, ``Crossmatch: Cross-view matching for semi-supervised remote sensing image segmentation,'' \emph{IEEE Transactions on Geoscience and Remote Sensing}, pp. 1--1, 2024.

\bibitem{chen2023self}
Y.~Chen, W.~Huang, S.~Zhou, Q.~Chen, and Z.~Xiong, ``Self-supervised neuron segmentation with multi-agent reinforcement learning,'' in \emph{IJCAI}, 2023, pp. 609--617.

\bibitem{chen2024learning}
Y.~Chen, W.~Huang, X.~Liu, S.~Deng, Q.~Chen, and Z.~Xiong, ``Learning multiscale consistency for self-supervised electron microscopy instance segmentation,'' in \emph{ICASSP}.\hskip 1em plus 0.5em minus 0.4em\relax IEEE, 2024, pp. 1566--1570.

\bibitem{qianmaskfactory}
H.~Qian, Y.~Chen, S.~Lou, F.~Khan, X.~Jin, and D.-P. Fan, ``Maskfactory: Towards high-quality synthetic data generation for dichotomous image segmentation,'' in \emph{NeurIPS}, 2024.

\bibitem{cha}
A.~Singh, ``Change detection in the tropical forest environment of northeastern india using landsat,'' \emph{Remote sensing and tropical land management}, vol.~44, pp. 273--254, 1986.

\bibitem{bi}
W.~J. Todd, ``Urban and regional land use change detected by using landsat data,'' \emph{Journal of Research of the US Geological Survey}, vol.~5, no.~5, pp. 529--534, 1977.

\bibitem{pca}
T.~Celik, ``Unsupervised change detection in satellite images using principal component analysis and $ k $-means clustering,'' \emph{IEEE geoscience and remote sensing letters}, vol.~6, no.~4, pp. 772--776, 2009.

\bibitem{cva}
S.~Saha, F.~Bovolo, and L.~Bruzzone, ``Unsupervised deep change vector analysis for multiple-change detection in vhr images,'' \emph{IEEE Transactions on Geoscience and Remote Sensing}, vol.~57, no.~6, pp. 3677--3693, 2019.

\bibitem{chen2024tokenunify}
Y.~Chen, H.~Shi, X.~Liu, T.~Shi, R.~Zhang, D.~Liu, Z.~Xiong, and F.~Wu, ``Tokenunify: Scalable autoregressive visual pre-training with mixture token prediction,'' \emph{arXiv preprint arXiv:2405.16847}, 2024.

\bibitem{chen2024bimcv}
Y.~Chen, C.~Liu, X.~Liu, R.~Arcucci, and Z.~Xiong, ``Bimcv-r: A landmark dataset for 3d ct text-image retrieval,'' in \emph{MICCAI}.\hskip 1em plus 0.5em minus 0.4em\relax Springer, 2024, pp. 124--134.

\bibitem{sunprogram}
H.~Sun, L.~Xu, S.~Jin, P.~Luo, C.~Qian, and W.~Liu, ``Program: Prototype graph model based pseudo-label learning for test-time adaptation,'' in \emph{The Twelfth International Conference on Learning Representations}.

\bibitem{cnn}
A.~Krizhevsky, I.~Sutskever, and G.~E. Hinton, ``Imagenet classification with deep convolutional neural networks,'' \emph{Communications of the ACM}, vol.~60, no.~6, pp. 84--90, 2017.

\bibitem{fc-siam}
R.~C. Daudt, B.~Le~Saux, and A.~Boulch, ``Fully convolutional siamese networks for change detection,'' in \emph{2018 25th IEEE International Conference on Image Processing (ICIP)}.\hskip 1em plus 0.5em minus 0.4em\relax IEEE, 2018, pp. 4063--4067.

\bibitem{vgg}
K.~Simonyan and A.~Zisserman, ``Very deep convolutional networks for large-scale image recognition,'' \emph{arXiv preprint arXiv:1409.1556}, 2014.

\bibitem{resnet}
K.~He, X.~Zhang, S.~Ren, and J.~Sun, ``Deep residual learning for image recognition,'' in \emph{Proceedings of the IEEE conference on computer vision and pattern recognition}, 2016, pp. 770--778.

\bibitem{ifnet}
C.~Zhang, P.~Yue, D.~Tapete, L.~Jiang, B.~Shangguan, L.~Huang, and G.~Liu, ``A deeply supervised image fusion network for change detection in high resolution bi-temporal remote sensing images,'' \emph{ISPRS Journal of Photogrammetry and Remote Sensing}, vol. 166, pp. 183--200, 2020.

\bibitem{stanet}
H.~Chen and Z.~Shi, ``A spatial-temporal attention-based method and a new dataset for remote sensing image change detection,'' \emph{Remote Sensing}, vol.~12, no.~10, p. 1662, 2020.

\bibitem{dtcdscn}
Y.~Liu, C.~Pang, Z.~Zhan, X.~Zhang, and X.~Yang, ``Building change detection for remote sensing images using a dual-task constrained deep siamese convolutional network model,'' \emph{IEEE Geoscience and Remote Sensing Letters}, vol.~18, no.~5, pp. 811--815, 2020.

\bibitem{tfigr}
Z.~Li, C.~Tang, L.~Wang, and A.~Y. Zomaya, ``Remote sensing change detection via temporal feature interaction and guided refinement,'' \emph{IEEE Transactions on Geoscience and Remote Sensing}, vol.~60, pp. 1--11, 2022.

\bibitem{snunet}
S.~Fang, K.~Li, J.~Shao, and Z.~Li, ``Snunet-cd: A densely connected siamese network for change detection of vhr images,'' \emph{IEEE Geoscience and Remote Sensing Letters}, vol.~19, pp. 1--5, 2021.

\bibitem{transformer}
A.~Vaswani, N.~Shazeer, N.~Parmar, J.~Uszkoreit, L.~Jones, A.~N. Gomez, {\L}.~Kaiser, and I.~Polosukhin, ``Attention is all you need,'' \emph{Advances in neural information processing systems}, vol.~30, 2017.

\bibitem{swinsunet}
C.~Zhang, L.~Wang, S.~Cheng, and Y.~Li, ``Swinsunet: Pure transformer network for remote sensing image change detection,'' \emph{IEEE Transactions on Geoscience and Remote Sensing}, vol.~60, pp. 1--13, 2022.

\bibitem{swin}
Z.~Liu, Y.~Lin, Y.~Cao, H.~Hu, Y.~Wei, Z.~Zhang, S.~Lin, and B.~Guo, ``Swin transformer: Hierarchical vision transformer using shifted windows,'' in \emph{Proceedings of the IEEE/CVF international conference on computer vision}, 2021, pp. 10\,012--10\,022.

\bibitem{changeformer}
W.~G.~C. Bandara and V.~M. Patel, ``A transformer-based siamese network for change detection,'' in \emph{IGARSS 2022-2022 IEEE International Geoscience and Remote Sensing Symposium}.\hskip 1em plus 0.5em minus 0.4em\relax IEEE, 2022, pp. 207--210.

\bibitem{bit}
H.~Chen, Z.~Qi, and Z.~Shi, ``Remote sensing image change detection with transformers,'' \emph{IEEE Transactions on Geoscience and Remote Sensing}, vol.~60, pp. 1--14, 2021.

\bibitem{vit}
A.~Dosovitskiy, L.~Beyer, A.~Kolesnikov, D.~Weissenborn, X.~Zhai, T.~Unterthiner, M.~Dehghani, M.~Minderer, G.~Heigold, S.~Gelly \emph{et~al.}, ``An image is worth 16x16 words: Transformers for image recognition at scale,'' \emph{arXiv preprint arXiv:2010.11929}, 2020.

\bibitem{ff1}
M.~Tancik, P.~Srinivasan, B.~Mildenhall, S.~Fridovich-Keil, N.~Raghavan, U.~Singhal, R.~Ramamoorthi, J.~Barron, and R.~Ng, ``Fourier features let networks learn high frequency functions in low dimensional domains,'' \emph{Advances in Neural Information Processing Systems}, vol.~33, pp. 7537--7547, 2020.

\bibitem{ff2}
B.~Zheng, S.~Yuan, C.~Yan, X.~Tian, J.~Zhang, Y.~Sun, L.~Liu, A.~Leonardis, and G.~Slabaugh, ``Learning frequency domain priors for image demoireing,'' \emph{IEEE Transactions on Pattern Analysis and Machine Intelligence}, vol.~44, no.~11, pp. 7705--7717, 2021.

\bibitem{fcanet}
Z.~Qin, P.~Zhang, F.~Wu, and X.~Li, ``Fcanet: Frequency channel attention networks,'' in \emph{Proceedings of the IEEE/CVF international conference on computer vision}, 2021, pp. 783--792.

\bibitem{hui}
Y.~Sun, L.~Lei, X.~Tan, D.~Guan, J.~Wu, and G.~Kuang, ``Structured graph based image regression for unsupervised multimodal change detection,'' \emph{ISPRS Journal of Photogrammetry and Remote Sensing}, vol. 185, pp. 16--31, 2022.

\bibitem{cap}
E.~P. Crist, ``A tm tasseled cap equivalent transformation for reflectance factor data,'' \emph{Remote sensing of Environment}, vol.~17, no.~3, pp. 301--306, 1985.

\bibitem{svm}
R.~G. Negri, A.~C. Frery, W.~Casaca, S.~Azevedo, M.~A. Dias, E.~A. Silva, and E.~H. Alc{\^a}ntara, ``Spectral--spatial-aware unsupervised change detection with stochastic distances and support vector machines,'' \emph{IEEE Transactions on Geoscience and Remote Sensing}, vol.~59, no.~4, pp. 2863--2876, 2020.

\bibitem{rand}
D.~K. Seo, Y.~H. Kim, Y.~D. Eo, M.~H. Lee, and W.~Y. Park, ``Fusion of sar and multispectral images using random forest regression for change detection,'' \emph{ISPRS International Journal of Geo-Information}, vol.~7, no.~10, p. 401, 2018.

\bibitem{spatial}
M.~Zhang, G.~Xu, K.~Chen, M.~Yan, and X.~Sun, ``Triplet-based semantic relation learning for aerial remote sensing image change detection,'' \emph{IEEE Geoscience and Remote Sensing Letters}, vol.~16, no.~2, pp. 266--270, 2018.

\bibitem{dminet}
Y.~Feng, J.~Jiang, H.~Xu, and J.~Zheng, ``Change detection on remote sensing images using dual-branch multilevel intertemporal network,'' \emph{IEEE Transactions on Geoscience and Remote Sensing}, vol.~61, pp. 1--15, 2023.

\bibitem{deeply1}
C.~Zhang, P.~Yue, D.~Tapete, L.~Jiang, B.~Shangguan, L.~Huang, and G.~Liu, ``A deeply supervised image fusion network for change detection in high resolution bi-temporal remote sensing images,'' \emph{ISPRS Journal of Photogrammetry and Remote Sensing}, vol. 166, pp. 183--200, 2020.

\bibitem{deeply2}
Q.~Shi, M.~Liu, S.~Li, X.~Liu, F.~Wang, and L.~Zhang, ``A deeply supervised attention metric-based network and an open aerial image dataset for remote sensing change detection,'' \emph{IEEE transactions on geoscience and remote sensing}, vol.~60, pp. 1--16, 2021.

\bibitem{mstdsnet}
F.~Song, S.~Zhang, T.~Lei, Y.~Song, and Z.~Peng, ``Mstdsnet-cd: Multiscale swin transformer and deeply supervised network for change detection of the fast-growing urban regions,'' \emph{IEEE Geoscience and Remote Sensing Letters}, vol.~19, pp. 1--5, 2022.

\bibitem{segformer}
E.~Xie, W.~Wang, Z.~Yu, A.~Anandkumar, J.~M. Alvarez, and P.~Luo, ``Segformer: Simple and efficient design for semantic segmentation with transformers,'' \emph{Advances in Neural Information Processing Systems}, vol.~34, pp. 12\,077--12\,090, 2021.

\bibitem{nie2024imputeformer}
T.~Nie, G.~Qin, W.~Ma, Y.~Mei, and J.~Sun, ``Imputeformer: Low rankness-induced transformers for generalizable spatiotemporal imputation,'' in \emph{Proceedings of the 30th ACM SIGKDD Conference on Knowledge Discovery and Data Mining}, 2024, pp. 2260--2271.

\bibitem{he2024geolocation}
J.~He, T.~Nie, and W.~Ma, ``Geolocation representation from large language models are generic enhancers for spatio-temporal learning,'' \emph{arXiv preprint arXiv:2408.12116}, 2024.

\bibitem{hivit}
X.~Zhang, Y.~Tian, L.~Xie, W.~Huang, Q.~Dai, Q.~Ye, and Q.~Tian, ``Hivit: A simpler and more efficient design of hierarchical vision transformer,'' in \emph{The Eleventh International Conference on Learning Representations}, 2023.

\bibitem{relative}
K.~Wu, H.~Peng, M.~Chen, J.~Fu, and H.~Chao, ``Rethinking and improving relative position encoding for vision transformer,'' in \emph{Proceedings of the IEEE/CVF International Conference on Computer Vision}, 2021, pp. 10\,033--10\,041.

\bibitem{refiner}
D.~Zhou, Y.~Shi, B.~Kang, W.~Yu, Z.~Jiang, Y.~Li, X.~Jin, Q.~Hou, and J.~Feng, ``Refiner: Refining self-attention for vision transformers,'' \emph{arXiv preprint arXiv:2106.03714}, 2021.

\bibitem{sun2024ultrahighresolutionsegmentationboundaryenhanced}
\BIBentryALTinterwordspacing
H.~Sun, ``Ultra-high resolution segmentation via boundary-enhanced patch-merging transformer,'' 2024. [Online]. Available: \url{https://arxiv.org/abs/2412.10181}
\BIBentrySTDinterwordspacing

\bibitem{cheng2024sptsequenceprompttransformer}
\BIBentryALTinterwordspacing
S.~Cheng and H.~Sun, ``Spt: Sequence prompt transformer for interactive image segmentation,'' 2024. [Online]. Available: \url{https://arxiv.org/abs/2412.10224}
\BIBentrySTDinterwordspacing

\bibitem{cvt}
H.~Wu, B.~Xiao, N.~Codella, M.~Liu, X.~Dai, L.~Yuan, and L.~Zhang, ``Cvt: Introducing convolutions to vision transformers,'' in \emph{Proceedings of the IEEE/CVF International Conference on Computer Vision}, 2021, pp. 22--31.

\bibitem{cmt}
J.~Guo, K.~Han, H.~Wu, Y.~Tang, X.~Chen, Y.~Wang, and C.~Xu, ``Cmt: Convolutional neural networks meet vision transformers,'' in \emph{Proceedings of the IEEE/CVF Conference on Computer Vision and Pattern Recognition}, 2022, pp. 12\,175--12\,185.

\bibitem{convit}
S.~d’Ascoli, H.~Touvron, M.~L. Leavitt, A.~S. Morcos, G.~Biroli, and L.~Sagun, ``Convit: Improving vision transformers with soft convolutional inductive biases,'' in \emph{International Conference on Machine Learning}.\hskip 1em plus 0.5em minus 0.4em\relax PMLR, 2021, pp. 2286--2296.

\bibitem{mlp}
I.~O. Tolstikhin, N.~Houlsby, A.~Kolesnikov, L.~Beyer, X.~Zhai, T.~Unterthiner, J.~Yung, A.~Steiner, D.~Keysers, J.~Uszkoreit \emph{et~al.}, ``Mlp-mixer: An all-mlp architecture for vision,'' \emph{Advances in neural information processing systems}, vol.~34, pp. 24\,261--24\,272, 2021.

\bibitem{mlps}
H.~Liu, Z.~Dai, D.~So, and Q.~V. Le, ``Pay attention to mlps,'' \emph{Advances in Neural Information Processing Systems}, vol.~34, pp. 9204--9215, 2021.

\bibitem{resmlp}
H.~Touvron, P.~Bojanowski, M.~Caron, M.~Cord, A.~El-Nouby, E.~Grave, G.~Izacard, A.~Joulin, G.~Synnaeve, J.~Verbeek \emph{et~al.}, ``Resmlp: Feedforward networks for image classification with data-efficient training,'' \emph{IEEE Transactions on Pattern Analysis and Machine Intelligence}, 2022.

\bibitem{poolormer}
W.~Yu, M.~Luo, P.~Zhou, C.~Si, Y.~Zhou, X.~Wang, J.~Feng, and S.~Yan, ``Metaformer is actually what you need for vision,'' in \emph{Proceedings of the IEEE/CVF conference on computer vision and pattern recognition}, 2022, pp. 10\,819--10\,829.

\bibitem{fnet}
J.~Lee-Thorp, J.~Ainslie, I.~Eckstein, and S.~Ontanon, ``Fnet: Mixing tokens with fourier transforms,'' \emph{arXiv preprint arXiv:2105.03824}, 2021.

\bibitem{asymmetric}
X.~Zhang, S.~Cheng, L.~Wang, and H.~Li, ``Asymmetric cross-attention hierarchical network based on cnn and transformer for bitemporal remote sensing images change detection,'' \emph{IEEE Transactions on Geoscience and Remote Sensing}, vol.~61, pp. 1--15, 2023.

\bibitem{ren2022shunted}
S.~Ren, D.~Zhou, S.~He, J.~Feng, and X.~Wang, ``Shunted self-attention via multi-scale token aggregation,'' in \emph{Proceedings of the IEEE/CVF conference on computer vision and pattern recognition}, 2022, pp. 10\,853--10\,862.

\bibitem{changestar}
Z.~Zheng, A.~Ma, L.~Zhang, and Y.~Zhong, ``Change is everywhere: Single-temporal supervised object change detection in remote sensing imagery,'' in \emph{Proceedings of the IEEE/CVF international conference on computer vision}, 2021, pp. 15\,193--15\,202.

\bibitem{dmatnet}
X.~Song, Z.~Hua, and J.~Li, ``Remote sensing image change detection transformer network based on dual-feature mixed attention,'' \emph{IEEE Transactions on Geoscience and Remote Sensing}, vol.~60, pp. 1--16, 2022.

\bibitem{lgpnet}
T.~Liu, M.~Gong, D.~Lu, Q.~Zhang, H.~Zheng, F.~Jiang, and M.~Zhang, ``Building change detection for vhr remote sensing images via local--global pyramid network and cross-task transfer learning strategy,'' \emph{IEEE Transactions on Geoscience and Remote Sensing}, vol.~60, pp. 1--17, 2022.

\bibitem{USSFCNet}
T.~Lei, X.~Geng, H.~Ning, Z.~Lv, M.~Gong, Y.~Jin, and A.~K. Nandi, ``Ultralightweight spatial--spectral feature cooperation network for change detection in remote sensing images,'' \emph{IEEE Transactions on Geoscience and Remote Sensing}, vol.~61, pp. 1--14, 2023.

\bibitem{group}
Y.~Ioannou, D.~Robertson, R.~Cipolla, and A.~Criminisi, ``Deep roots: Improving cnn efficiency with hierarchical filter groups,'' in \emph{Proceedings of the IEEE conference on computer vision and pattern recognition}, 2017, pp. 1231--1240.

\bibitem{focal}
T.-Y. Lin, P.~Goyal, R.~Girshick, K.~He, and P.~Doll{\'a}r, ``Focal loss for dense object detection,'' in \emph{Proceedings of the IEEE international conference on computer vision}, 2017, pp. 2980--2988.

\bibitem{dice}
X.~Li, X.~Sun, Y.~Meng, J.~Liang, F.~Wu, and J.~Li, ``Dice loss for data-imbalanced nlp tasks,'' \emph{arXiv preprint arXiv:1911.02855}, 2019.

\bibitem{adam}
D.~P. Kingma and J.~Ba, ``Adam: A method for stochastic optimization,'' \emph{arXiv preprint arXiv:1412.6980}, 2014.

\bibitem{whu}
S.~Ji, S.~Wei, and M.~Lu, ``Fully convolutional networks for multisource building extraction from an open aerial and satellite imagery data set,'' \emph{IEEE Transactions on Geoscience and Remote Sensing}, vol.~57, no.~1, pp. 574--586, 2018.

\bibitem{levir}
H.~Chen and Z.~Shi, ``A spatial-temporal attention-based method and a new dataset for remote sensing image change detection,'' \emph{Remote Sensing}, vol.~12, no.~10, p. 1662, 2020.

\bibitem{clcd}
M.~Liu, Z.~Chai, H.~Deng, and R.~Liu, ``A cnn-transformer network with multi-scale context aggregation for fine-grained cropland change detection,'' \emph{IEEE Journal of Selected Topics in Applied Earth Observations and Remote Sensing}, 2022.

\end{thebibliography}










\end{document}